%% file: main.tex
\documentclass[10pt,twocolumn,letterpaper]{article}

\usepackage[pagenumbers]{cvpr} 
\usepackage{graphicx}
\usepackage{xcolor}
\usepackage{multirow}
\usepackage{comment}

\usepackage{booktabs}
\usepackage{xcolor}

\input{preamble}
\definecolor{cvprblue}{rgb}{0.21,0.49,0.74}
\usepackage[pagebackref,breaklinks,colorlinks,allcolors=cvprblue]{hyperref}
\definecolor{cClipB32}{RGB}{0, 130, 130}       
\definecolor{cClipB16}{RGB}{220, 105, 0}       
\definecolor{cDino}{RGB}{125, 40, 140}         
\definecolor{cFloor32}{RGB}{100, 100, 100}     
\definecolor{cFloor16}{RGB}{40, 90, 180}       

\usepackage[margin=1in]{geometry}
\usepackage{tikz}
\usepackage{subcaption}
\usetikzlibrary{shapes.geometric, positioning, matrix}
\usepackage{amsmath}

\def\confName{CVPR}
\def\confYear{2027}

\title{Towards Closing the Performance Gap in Cross-Modal Place Recognition
}

\author{Ofer Idan \qquad Dan Badur \qquad Yoli Shavit\thanks{Corresponding author.}\\
Faculty of Engineering, Bar-Ilan University, Ramat Gan, Israel\\
{\tt\small \{ofer.idan, badur.dan, yoli.shavit\}@biu.ac.il}
}

\begin{document}
\maketitle
\input{sec/0_abstract}    
\input{sec/1_intro}

\input{sec/2_related_work}
\input{sec/3_method}
\input{sec/4_experimental_setup}

\input{sec/5_results}
\input{sec/6_conclusions}



{
    \small
    \bibliographystyle{ieeenat_fullname}
    \bibliography{main}
}

\section*{Supplementary Material}
\setcounter{section}{0}
\renewcommand{\thesection}{\Alph{section}}

\input{sec/X_suppl}
\end{document}

%% file: sec/0_abstract.tex
\begin{abstract}
Cross-modal place recognition (CMPR) matches natural language descriptions against databases of geo-tagged reference images, enabling localization without visual queries. Despite its potential for language-guided navigation, existing vision-language models lag far behind visual place recognition (VPR) baselines, even after task adaptation. 
Unlike general-purpose text-to-image retrieval, place recognition relies on subtle local cues to distinguish visually similar locations. We therefore hypothesize that a primary bottleneck in CMPR is capturing fine-grained correspondences between image regions and text details, which requires higher visual token granularity, strong text representation capacity, and local cross-modal matching.
To address this challenge, we propose CrossPlace, a two-stage CMPR framework. Stage 1 performs global retrieval by pairing a jointly aligned DINOv2-B/14 visual backbone with a progressively fine-tuned BGE-Large text encoder, while Stage 2 refines top candidates using a lightweight token-level cross-attention reranker. Through systematic analysis, we isolate the drivers of retrieval accuracy and show that visual token density provides the largest performance improvement, while text encoder capacity and task fine-tuning offer complementary gains. To evaluate robustness under realistic conditions, we also extend CMPR evaluation to city-scale and severe weather and illumination settings. Across all existing and new benchmarks, CrossPlace achieves double-digit Recall@1 improvements over prior cross-modal methods, closes the gap to unimodal text retrieval, and narrows it to unimodal VPR. 
Code and benchmarks are available at \url{https://github.com/oferidan1/CrossPlace}
\end{abstract}

%% file: sec/1_intro.tex
\section{Introduction}
\label{sec:intro}

Visual place recognition (VPR) matches visual queries against geo-localized image databases for global relocalization across diverse applications~\cite{torii2013visual,lu2024cricavpr}. Cross-modal place recognition (CMPR), introduced by LaVPR~\cite{idan2026lavpr}, extends this task to natural language, retrieving geographic locations purely from textual descriptions without an input image.

CMPR addresses a fundamental operational gap. For example, in search-and-rescue operations, first responders verbally describe observed landmarks that must be geolocated immediately. Similarly, human-in-the-loop navigation requires operators to direct autonomous systems through language when camera feeds are unavailable. More broadly, text queries offer unique flexibility when image acquisition is corrupted, physically impossible, or operationally impractical, conditions that fundamentally limit or break standard VPR pipelines.

\input{figures/teaser}

Despite the promise of cross-modal localization, language-to-image retrieval substantially trails unimodal visual place recognition. For example, on the MSLS-Val benchmark~\cite{msls}, the best-performing cross-modal method (LaVPR~\cite{idan2026lavpr}) achieves a Recall@1 of $38.0\%$, whereas state-of-the-art vision-only models exceed $85\%$~\cite{lu2024cricavpr,izquierdo2024optimal}.
Unlike general-purpose text-to-image retrieval, which focuses on coarse semantic categories, place recognition distinguishes specific locations through subtle local cues, such as door styles, window shapes, and facade textures, and their spatial layout within the scene. 

We propose \textit{CrossPlace}, a two-stage CMPR framework designed to address this challenge through higher visual token granularity, strong text representation capacity, and local cross-modal matching. In Stage~1, CrossPlace performs global candidate retrieval by pairing a jointly aligned DINOv2-B/14 visual backbone with a progressively fine-tuned BGE-Large text encoder. In Stage~2, inspired by spatial reranking in visual place recognition~\cite{tzachor2025effovpr}, we employ a lightweight token-level cross-attention reranker over text tokens and local image patches to refine candidate scores.

Through systematic analysis of CrossPlace and alternative configurations, we isolate the drivers of retrieval accuracy in practice. Our investigation shows that visual token density (spatial patch granularity) is the primary source of performance gains: increasing patch density from $49$ tokens in ViT-B/32 to $256$ in DINOv2-B/14 yields the largest performance improvement across all benchmarks, further reinforced by self-supervised spatial representations. Meanwhile, text encoder capacity and progressive task fine-tuning provide complementary, supporting gains once sufficient visual granularity is established. Furthermore, because CrossPlace matches or closely approaches the unimodal text-to-text retrieval reference (\cref{fig:teaser}, dotted cyan line), these findings suggest that global cross-modal alignment is a secondary bottleneck.

To establish a broader benchmark suite and assess scalability and domain generalization, we evaluate our approach on established LaVPR benchmarks and introduce new cross-modal benchmarks derived from San Francisco eXtra Large (SF-XL) and SVOX. Across these diverse settings, our global retrieval stage alone substantially outperforms existing baselines, improving Recall@1 by $5.2\text{--}20.3$ percentage points (pp). Incorporating our lightweight cross-attention reranking module yields an overall improvement of $9.7\text{--}32.4$~pp over prior cross-modal state of the art across all benchmarks, while introducing negligible latency. Moreover, CrossPlace narrows the gap to unimodal VPR baselines (\cref{fig:teaser}, dashed orange line).

In summary, our key contributions are as follows:
\begin{itemize}
    \item We propose \textit{CrossPlace}, a two-stage CMPR framework designed to capture fine-grained image--text correspondences, pairing a jointly aligned DINOv2-B/14 visual backbone with a progressively fine-tuned BGE-Large text encoder in Stage~1, followed by a lightweight token-level cross-attention reranker in Stage~2.
    \item Through systematic architectural analysis, we pinpoint the drivers of CMPR accuracy, showing that visual token density is the primary source of performance improvement, while text encoder capacity and progressive task alignment provide complementary gains, effectively closing the gap to unimodal text retrieval.
    \item We extend CMPR evaluation to city-scale (SF-XL) and severe weather and illumination (SVOX) settings, demonstrating that CrossPlace achieves double-digit Recall@1 improvements over prior cross-modal baselines across all existing and new benchmarks.
\end{itemize}

%% file: figures/teaser.tex
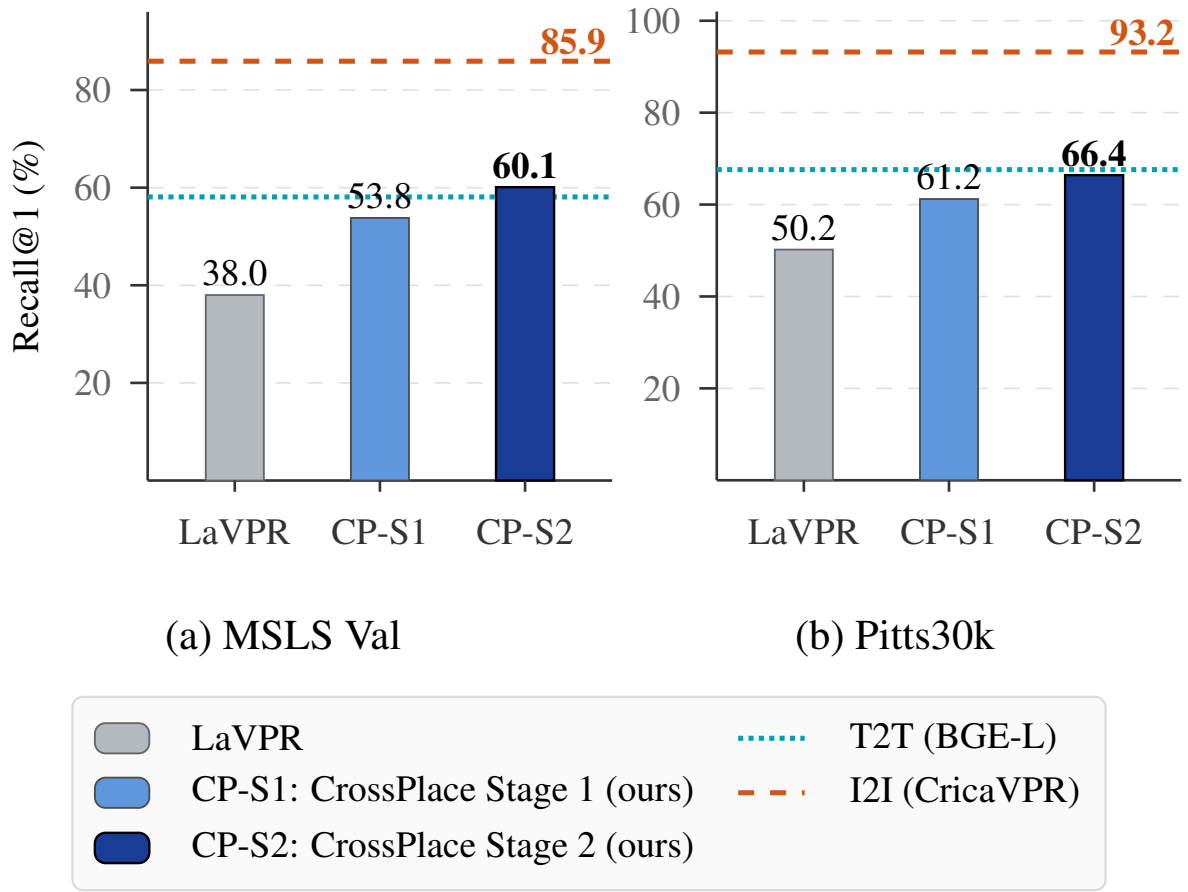
\begin{figure}[t]
\centering

\definecolor{cBlip}{RGB}{180, 185, 190}
\definecolor{cStage1}{RGB}{95, 150, 220}
\definecolor{cStage2}{RGB}{25, 60, 150}
\definecolor{cT2T}{RGB}{0, 160, 180}
\definecolor{cI2I}{RGB}{210, 85, 20}

\tikzset{
    plotgrid/.style={gray!25, dashed, line width=0.3pt},
    plotaxis/.style={black!80, line width=0.5pt},
    bT2T/.style={cT2T, densely dotted, line width=0.95pt},
    bI2I/.style={cI2I, dashed, line width=0.95pt},
    barBlip/.style={fill=cBlip, draw=black!60, line width=0.35pt},
    barStage1/.style={fill=cStage1, draw=black!70, line width=0.35pt},
    barStage2/.style={fill=cStage2, draw=black, line width=0.45pt}
}

\begin{subfigure}[b]{0.485\linewidth}
\centering
\begin{tikzpicture}[x=0.96cm, y=0.03229cm]
    \foreach \y in {20,40,60,80} {
        \draw[plotgrid] (0.4,\y) -- (3.6,\y);
        \draw[plotaxis] (0.4,\y) -- (0.28,\y)
            node[left, font=\scriptsize, text=gray!80!black] {\y};
    }
    \node[rotate=90, font=\scriptsize] at (-0.42, 48) {Recall@1 (\%)};
    \draw[plotaxis] (0.4,0) -- (0.4,96);
    \draw[plotaxis] (0.4,0) -- (3.6,0);
    
    \draw[bT2T] (0.4,58.1) -- (3.6,58.1);
    \draw[bI2I] (0.4,85.9) -- (3.6,85.9);
    \node[above left, font=\scriptsize\bfseries, text=cI2I, inner sep=1pt] at (3.6,85.9) {85.9};

    \draw[barBlip]   (0.8,0) rectangle (1.2,38.0);
    \node[above,font=\scriptsize, inner sep=1.5pt] at (1.0,38.0) {38.0};
    \draw[barStage1] (1.8,0) rectangle (2.2,53.8);
    \node[above,font=\scriptsize, inner sep=1.5pt] at (2.0,53.8) {53.8};
    \draw[barStage2] (2.8,0) rectangle (3.2,60.1);
    \node[above,font=\scriptsize\bfseries, inner sep=1.5pt] at (3.0,60.1) {60.1};
    
    \foreach \x/\lbl in {1.0/{LaVPR}, 2.0/{CP-S1}, 3.0/{CP-S2}} {
        \draw[plotaxis] (\x,0) -- (\x,-2.3)
            node[below=1.5pt, font=\scriptsize] {\lbl};
    }
\end{tikzpicture}
\vspace{-1.2ex}
\caption{MSLS Val}
\label{fig:vpr_msls}
\end{subfigure}
\hfill
\begin{subfigure}[b]{0.485\linewidth}
\centering
\begin{tikzpicture}[x=0.96cm, y=0.03039cm]
    \foreach \y in {20,40,60,80,100} {
        \draw[plotgrid] (0.4,\y) -- (3.6,\y);
        \draw[plotaxis] (0.4,\y) -- (0.28,\y)
            node[left, font=\scriptsize, text=gray!80!black] {\y};
    }
    \draw[plotaxis] (0.4,0) -- (0.4,102);
    \draw[plotaxis] (0.4,0) -- (3.6,0);
    
    \draw[bT2T] (0.4,67.6) -- (3.6,67.6);
    \draw[bI2I] (0.4,93.2) -- (3.6,93.2);
    \node[above left, font=\scriptsize\bfseries, text=cI2I, inner sep=1pt] at (3.6,93.2) {93.2};

    \draw[barBlip]   (0.8,0) rectangle (1.2,50.2);
    \node[above,font=\scriptsize, inner sep=1.5pt] at (1.0,50.2) {50.2};
    \draw[barStage1] (1.8,0) rectangle (2.2,61.2);
    \node[above,font=\scriptsize, inner sep=1.5pt] at (2.0,61.2) {61.2};
    \draw[barStage2] (2.8,0) rectangle (3.2,66.4);
    \node[above,font=\scriptsize\bfseries, inner sep=1.5pt] at (3.0,66.4) {66.4};
    
    \foreach \x/\lbl in {1.0/{LaVPR}, 2.0/{CP-S1}, 3.0/{CP-S2}} {
        \draw[plotaxis] (\x,0) -- (\x,-2.5)
            node[below=1.5pt, font=\scriptsize] {\lbl};
    }
\end{tikzpicture}
\vspace{-1.2ex}
\caption{Pitts30k}
\label{fig:vpr_pitts}
\end{subfigure}

\vspace{0.5em}
\begin{tikzpicture}[font=\scriptsize]
    \matrix[
        matrix of nodes,
        nodes={anchor=west, inner sep=0.8pt},
        column sep=0.7em,
        row sep=0.25em,
        draw=gray!30,
        rounded corners=2pt,
        inner sep=4pt,
        fill=gray!5
    ] {
        \draw[barBlip]   (-0.18,-0.1) rectangle (0.18,0.1); & \node{LaVPR}; &
        \draw[bT2T] (-0.22,0) -- (0.22,0); & \node{T2T (BGE-L)}; \\
        \draw[barStage1] (-0.18,-0.1) rectangle (0.18,0.1); & \node{CP-S1: CrossPlace Stage 1 (ours)}; &
        \draw[bI2I] (-0.22,0) -- (0.22,0); & \node{I2I (CricaVPR)}; \\
        \draw[barStage2] (-0.18,-0.1) rectangle (0.18,0.1); & \node{CP-S2: CrossPlace Stage 2 (ours)}; &
        & \\
    };
\end{tikzpicture}

\caption{Cross-modal place recognition performance (Recall@1) on the MSLS Val and Pitts30k benchmarks. Bars depict cross-modal models: LaVPR denotes the LaVPR cross-modal model (its best-performing variant, using BLIP). CrossPlace Stage~1 (CP-S1) is our global retrieval model, and CrossPlace Stage~2 (CP-S2) adds cross-attention reranking. Horizontal reference lines contextualize cross-modal retrieval relative to unimodal performance: the dotted cyan line marks unimodal text-to-text retrieval using BGE-Large, while the dashed orange line gives the performance of visual (image-to-image) place recognition using CricaVPR.}
\label{fig:teaser}
\end{figure}

%% file: sec/2_related_work.tex
\section{Related Work}
\label{sec:related_work}

\subsection{Visual Place Recognition}
\label{subsec:rel_vpr}
\paragraph{Image-based Place Recognition.}
Visual place recognition (VPR) identifies the location of a query image by retrieving nearest matches from a geo-tagged database. Global descriptors are typically formed via pooling layers such as NetVLAD~\cite{arandjelovic2016netvlad} or GeM~\cite{radenovic2018fine}, with recent efforts boosting discriminability through optimal transport~\cite{izquierdo2024optimal}, feature fusion~\cite{ali2023mixvpr, lu2024cricavpr} and foundation model adaptation~\cite{lu2024towards, tzachor2025effovpr}. Training objectives primarily rely on the contrastive multi-similarity loss~\cite{Alibey_2022_gsvcities, wang2019multi} to enforce condition invariance, or classification proxies~\cite{berton2022rethinking, berton2023eigenplaces}.
Despite their effectiveness, image-only place recognition fails whenever query images cannot be acquired due to sensor failure, severe visual degradation~\cite{msls,torii2013visual}, or human-in-the-loop operation, motivating cross-modal paradigms that localize places directly from descriptive language queries~\cite{idan2026lavpr}.

\subsection{Cross-Modal Place Recognition (CMPR)}
\label{subsec:rel_cross_modal}
While standard visual localization typically operates in a unimodal mode, related efforts explore heterogeneous inputs. One line of work examines multimodal fusion, combining 2D imagery with 3D LiDAR point clouds~\cite{piasco2021improving,komorowski2021minkloc++,lai2022adafusion,hu2020dasgil}, or augmenting visual queries with extracted scene text~\cite{hong2019textplace}, descriptive language~\cite{idan2026lavpr}, and multi-sensor streams~\cite{melekhin2025mssplace}. Unified architectures like UniLoc~\cite{xia2024uniloc} further support diverse query modalities mapped against 3D submaps. Other formulations address text-to-3D localization~\cite{kolmet2022text2pos,xia2024text2loc,liu2025text,xu2025cmmloc} by mapping descriptions directly to 3D point cloud coordinates, or match text against overhead satellite and map tiles~\cite{ye2025cross}, diverging from scalable street-level visual databases. Recent cross-modal geo-localization methods also explore multi-anchor projection similarity for geometric alignment~\cite{hu2026maps}, multi-view text-vision registration~\cite{shang2025bridging}, and semantic-anchored multi-view foundation models bridging images and text for geo-localization~\cite{song2026geobridge}. These formulations rely on multi-view imagery or specialized geometric anchors, and do not apply directly to standard text-to-image place recognition or its established benchmarks~\cite{idan2026lavpr}.

Our work focuses on cross-modal text-to-image place recognition, where geographic locations in a street-level image database are retrieved purely from an input natural language description. LaVPR~\cite{idan2026lavpr} formalized this setting, showing that adapting vision-language models via LoRA~\cite{hu2022lora} with Multi-Similarity loss~\cite{wang2019multi} substantially outperforms zero-shot baselines. However, even after task-specific adaptation these models trail VPR baselines. In this work, we aim to narrow this gap by introducing CrossPlace.

\subsection{Reranking in Place Recognition}
\label{subsec:rel_reranking}
Two-stage coarse-to-fine pipelines are well established in VPR. Retrieval precision beyond global descriptors is typically enhanced through geometric verification~\cite{sarlin2019coarse}, local patch matching~\cite{hausler2021patch}, learned keypoint correspondence networks~\cite{sarlin2020superglue}, or learned candidate re-scorers~\cite{zhu2023r2former}. More recently, foundation model approaches like SelaVPR++~\cite{lu2025selavpr++} and EffoVPR~\cite{tzachor2025effovpr} demonstrate that adapting token-level features from frozen backbones yields highly efficient two-stage visual retrieval. However, all existing place recognition rerankers operate exclusively on image-to-image correspondences, comparing visual query features directly to database visual representations.

In contrast, two-stage reranking remains completely unexplored in cross-modal place recognition, where existing formulations like LaVPR~\cite{idan2026lavpr} rely strictly on single-stage global descriptor retrieval. While fine-grained late-interaction has been explored in general vision-and-language domains~\cite{messina2021fine,colbert}, those formulations do not transfer straightforwardly to cross-modal place recognition. General text-to-image retrieval methods like TERAN~\cite{messina2021fine} align words to salient object proposals produced by pre-trained detectors. Such object-centric proposals are computationally prohibitive across city-scale databases and discard the background architectural context and structural textures essential for disambiguating places. 
To bridge this gap, we introduce the first lightweight cross-attention reranker tailored for language-to-vision place recognition. Our architecture refines candidate similarity by attending text tokens directly to precomputed visual image patches, capturing fine-grained spatial correspondences with modest latency. 

%% file: sec/3_method.tex
\section{Method}
\label{sec:method}

We formulate the cross-modal place recognition problem (\Cref{subsec:problem_formulation}), detail our Stage~1 global retrieval architecture and its alignment scheme (\Cref{subsec:dense_retrieval}), and present our lightweight token-to-patch reranking module (\Cref{subsec:reranking}).

\subsection{Problem Formulation}
\label{subsec:problem_formulation}

Let $\mathcal{D} = \{I_i\}_{i=1}^{N}$ denote a database of geo-tagged images and $t_q$ a natural language query describing a target location. The cross-modal place recognition task retrieves the database image $I^* \in \mathcal{D}$ that corresponds to the described location. In a coarse-to-fine framework, retrieval begins by projecting the text query and database images into a shared metric space to select a shortlist of the top-$K$ candidate images based on global embedding similarity (e.g., cosine similarity). The shortlist is then reranked using fine-grained token-level correspondences to determine the final prediction.

\subsection{Cross-Modal Global Retrieval (Stage 1)}
\label{subsec:dense_retrieval}

\paragraph{Backbone Formulation.}
Let $f_v$ denote the visual backbone that extracts a global representation $f_v(I) \in \mathbb{R}^{d_v}$ from an input image $I$, and let $f_t$ denote the text encoder that maps an input description $t$ to a pooled sentence embedding $f_t(t) \in \mathbb{R}^{d_t}$. Both encoders are adapted using Low-Rank Adaptation (LoRA)~\cite{hu2022lora} applied to all linear projection layers with rank $r$ following LaVPR~\cite{idan2026lavpr}. Learned linear heads ${h}_v: \mathbb{R}^{d_v} \to \mathbb{R}^{d_c}$ and ${h}_t: \mathbb{R}^{d_t} \to \mathbb{R}^{d_c}$ map the respective pooled representations into a shared $d_c$-dimensional metric space:
\begin{equation}
\mathbf{z}_v = {h}_v(f_v(I)), \quad \mathbf{z}_t = {h}_t(f_t(t)), \quad \mathbf{z}_v, \mathbf{z}_t \in \mathbb{R}^{d_c}
\label{eq:descriptors}
\end{equation}
Both descriptors are L2-normalised. Global similarity between query $t_q$ and candidate image $I_i$ is evaluated via cosine similarity $S(\mathbf{z}_t, \mathbf{z}_v^i) = \mathbf{z}_t^\top \mathbf{z}_v^i$. In our implementation, we instantiate the visual backbone $f_v$ with DINOv2-B~\cite{oquab2023dinov2} ($d_v = 768$), the text encoder $f_t$ with BGE-Large~\cite{bge_embedding} ($d_t = 1024$), and set $r = 64$ and $d_c = 768$.

\paragraph{Alignment and Training Objectives.}
Directly optimizing cross-modal alignment from generic pretrained checkpoints can cause representations to drift, because the text encoder must simultaneously adapt to domain-specific place descriptions and align with visual features. To prevent this, we adopt a two-stage training strategy governed by a general Multi-Similarity (MS) loss formulation~\cite{wang2019multi}. For a batch of queries $\mathcal{B}$, the loss over pairwise cosine similarities $S_{qp}$ and $S_{qn}$ is defined as
\begin{equation}\label{eq:ms_loss}
\begin{aligned}
\mathcal{L}_{\text{MS}} = \frac{1}{|\mathcal{B}|} \sum_{q \in \mathcal{B}} \Bigg\{ 
&\frac{1}{\gamma_p} \log\left[1 + \sum_{p \in \mathcal{P}_q} e^{-\gamma_p(S_{qp} - \lambda)}\right] \\
+ &\frac{1}{\gamma_n} \log\left[1 + \sum_{n \in \mathcal{N}_q} e^{\gamma_n(S_{qn} - \lambda)}\right] \Bigg\}
\end{aligned}
\end{equation}
where $\mathcal{P}_q$ and $\mathcal{N}_q$ denote the mined positive and negative candidate sets for query $q$, and $\gamma_p$, $\gamma_n$, and $\lambda$ are scaling hyperparameters.

We instantiate \cref{eq:ms_loss} across two successive training phases:
\begin{enumerate}
    \item \textit{Text-to-Text Domain Adaptation.} We first align the text encoder exclusively on text-to-text retrieval. Given paired descriptions of the same geographic location viewed from different viewpoints, we optimize BGE-Large via LoRA where $S_{qp}$ and $S_{qn}$ represent pairwise text-to-text similarities $S^{\text{tt}} = S(\mathbf{z}_t^q, \mathbf{z}_t^i)$. This unimodal stage establishes strong linguistic representations tailored specifically to place discrimination.
    \item \textit{Joint Cross-Modal Alignment.} Using the text encoder weights from the first step as initialization, we then jointly train the adapted BGE-Large encoder and the DINOv2 vision backbone. Here, the similarities in \cref{eq:ms_loss} become cross-modal text-to-vision similarities $S^{\text{tv}} = S(\mathbf{z}_t^q, \mathbf{z}_v^i)$. This joint alignment anchors the shared metric space, enabling the model to retrieve a candidate shortlist of top-$K$ database images for subsequent reranking.
\end{enumerate}

\subsection{Cross-Modal Reranking (Stage 2)}
\label{subsec:reranking}

\paragraph{Token-to-Patch Cross-Attention.}
While Stage~1 retrieval maps an image into a single global vector, queries frequently describe multiple specific spatial landmarks. To resolve fine-grained ambiguities, our Stage~2 module computes interactions between language tokens and local visual regions.

For a query $t_q$, we extract unpooled text token representations $\mathbf{T}_q \in \mathbb{R}^{L \times d_t}$ from the Stage~1 text encoder, where $L$ is the token sequence length. For candidate image $I_i$, we extract unpooled visual tokens $\mathbf{P}_i \in \mathbb{R}^{N_p \times d_v}$ from the Stage~1 visual backbone, where $N_p$ is the number of local patches. To extract local features efficiently without backpropagating into the heavy backbones, we merge the LoRA adapters from Stage~1 into their respective base models and keep all backbone parameters frozen during reranker training.

Because both encoders are optimized under global pooling objectives, their local token spaces lack elementwise alignment, which prevents the effective use of parameter-free late-interaction operators. We therefore introduce a lightweight learned cross-attention module. The text and visual tokens are linearly projected to a shared inner dimension $d$ to form matrices $\mathbf{X}_0 \in \mathbb{R}^{L \times d}$ and $\mathbf{P} \in \mathbb{R}^{N_p \times d}$, respectively. The sequence is then processed through $n_\ell$ interaction blocks. Each block applies cross-attention from text tokens to visual patches, followed by text self-attention and a feed-forward network:
\begin{align}
    \mathbf{X}_\ell' &= \mathbf{X}_{\ell-1} + \text{MHA}(\text{LN}(\mathbf{X}_{\ell-1}),\, \text{LN}(\mathbf{P}),\, \text{LN}(\mathbf{P})), \label{eq:cross_attn}\\
    \mathbf{X}_\ell'' &= \mathbf{X}_\ell' + \text{MHA}(\text{LN}(\mathbf{X}_\ell'),\, \text{LN}(\mathbf{X}_\ell'),\, \text{LN}(\mathbf{X}_\ell')), \label{eq:self_attn}\\
    \mathbf{X}_\ell &= \mathbf{X}_\ell'' + \text{FFN}(\text{LN}(\mathbf{X}_\ell'')) \label{eq:ffn}
\end{align}
where $\text{LN}$ denotes Layer Normalization, $\text{MHA}$ denotes multi-head attention with $h$ heads, and $\ell \in \{1, \dots, n_\ell\}$. After $n_\ell$ blocks, the contextualized text CLS token is passed through a linear projection vector $\mathbf{w} \in \mathbb{R}^d$ to produce a scalar reranking score:
\begin{equation}
    R(t_q, I_i) = \mathbf{w}^\top \mathrm{LN}\!\left(\mathbf{x}_{\text{CLS}}^{(n_\ell)}\right).
    \label{eq:score}
\end{equation}
At inference time, the final match score combines both stages linearly as:
\begin{equation}
    \hat{S}(t_q, I_i) = \alpha\, S(t_q, I_i) + (1-\alpha)\, R(t_q, I_i)
    \label{eq:score_fusion}
\end{equation}
where $S(t_q, I_i)$ and $R(t_q, I_i)$ denote the scores from Stage~1 retrieval (cosine similarity) and Stage~2 reranking (\cref{eq:score}), respectively, and $\alpha$ is a weighting coefficient.
 
\paragraph{Reranking Objective.}
The reranker is trained using hard negatives mined offline via the Stage~1 global retriever. For each training query $q$, Stage~1 retrieval produces a ranked candidate pool from which we sample a training shortlist containing at least one positive match. We optimize the reranker using a listwise cross-entropy objective over the candidate shortlist:
\begin{equation}
    \mathcal{L}_{\text{listwise}} = -\frac{1}{|\mathcal{B}|} \sum_{q \in \mathcal{B}} \sum_{k \in \mathcal{V}_q} \tilde{y}_k \log \frac{\exp(R_k)}{\sum_{j \in \mathcal{V}_q} \exp(r_j)},
    \label{eq:listwise}
\end{equation}
where $\mathcal{B}$ denotes the training batch, $\mathcal{V}_q$ is the set of shortlist candidate indices for query $q$, and $R_k = R(t_q, I_k)$ is the predicted reranking score for candidate $k$. The normalized target distribution $\tilde{y}_k = y_k / \sum_{j \in \mathcal{V}_q} y_j$ assigns uniform probability mass across all positive candidates ($y_k \in \{0, 1\}$) in $\mathcal{V}_q$ and zero to negatives.

We provide an extended description of our method, along with a visual overview of its inference pipeline, in the supplementary (suppl.) material.

%% file: sec/4_experimental_setup.tex
\section{Experimental Setup}
\label{sec:impl}

\textbf{Datasets.}
We evaluate CrossPlace on the cross-modal place recognition (CMPR) benchmarks introduced by LaVPR~\cite{idan2026lavpr}, which pair the  Pittsburgh30k (Pitts30k)~\cite{torii2013visual}, MSLS-val~\cite{msls}, and AmsterTime~\cite{amstertime} datasets with textual query descriptions. To establish a broader evaluation suite, we further apply the description-generation pipeline, available from LaVPR,  to two challenging benchmarks: SVOX~\cite{SVOX}, spanning severe weather and illumination shifts, and San Francisco eXtra Large (SF-XL)~\cite{berton2022rethinking}, a large-scale urban dataset. 

\paragraph{Baselines.}
We compare CrossPlace against the prior cross-modal state of the art (LaVPR), adopting LoRA-MS-BLIP~\cite{idan2026lavpr} as our primary baseline. In the systematic LaVPR evaluation, which spans zero-shot and PEFT-tuned variants of CLIP, BLIP, SigLIP2, and EVA-CLIP2, LoRA-MS-BLIP achieved the strongest CMPR performance. All fine-tuned LaVPR variants apply LoRA ($r{=}64$) across all linear projections, matching our Stage~1 setup and ensuring that our comparison isolates backbone architecture rather than the adaptation recipe. We evaluate LaVPR using the authors' publicly released code and checkpoints under our unified protocol. Results on standard benchmarks reproduce those reported in~\cite{idan2026lavpr}, and the same checkpoint is evaluated directly on SVOX and SF-XL.
\vspace{-1em}
\paragraph{Evaluation Protocol.}
We report Recall@$k$ (R@$k$) for $k \in \{1, 5, 10\}$, following
the standard VPR evaluation protocol~\cite{msls}. A retrieved image
is considered a correct match if it lies within 25\,m of the query
location for Pittsburgh30k, MSLS, and SF-XL, or shares the same
place identity for AmsterTime and SVOX. Stage~1 R@$k$ refers to
global embedding retrieval, while Stage~2 R@$k$ reports results after
cross-modal reranking of the top-$K$ shortlist.
\vspace{-1em}
\paragraph{Training and Implementation Details.}
For Stage~1 training, we use
batches of 80 places with 4 images each (320 images per
batch) and the MS loss (\cref{eq:ms_loss}) with
$\gamma_p{=}2$, $\gamma_n{=}40$, $\lambda{=}0.5$. Optimization
uses AdamW with initial learning rate $2{\times}10^{-5}$, decayed
at epochs 10 and 16, for 20 epochs total. LoRA
($r{=}64$) is applied to all linear projections of both BGE-Large
and DINOv2-B. For DINOv2-B, this covers the query, key, value,
and output projections and the MLP layers across all 12 transformer
blocks. DINOv2-B processes images at $224{\times}224$ resolution,
yielding $N_p{=}256$ patch tokens of dimension 768 with patch
size 14. BGE-Large tokenises descriptions with a maximum sequence length of 512 tokens (covering the maximal length of LaVPR descriptions).
All experiments are conducted on a single RTX Pro 6000 Blackwell
GPU with 96\,GB VRAM. 

For Stage~2, local patch and text tokens are precomputed from
the frozen Stage~1 encoders (where the LoRA matrices are included in the frozen encoder) and stored on disk, so
the backbone never runs at reranker training time. We use $d{=}128$,
$h{=}4$ attention heads, $n_\ell{=}1$ layers, batch size 64, AdamW
with learning rate $10^{-4}$ and cosine decay for 10 epochs.  The candidate pool for the Listwise loss (\cref{eq:listwise}) is constructed by taking the top-$K_{\text{cand}} = 50$ entries from a Stage~1 shortlist of 100 images per query.
Score fusion (\cref{eq:score_fusion}) uses $\alpha{=}0.2$, selected by ablation on Pitts30k validation set (see \cref{fig:alpha_ablation}).

All cross-modal retrieval models are trained on GSV-Cities~\cite{Alibey_2022_gsvcities} ($529{,}506$ images across $67{,}331$ places, with four views per place) augmented with LaVPR textual descriptions~\cite{idan2026lavpr}, and evaluated using the standard VPR protocol adopted in prior work~\cite{idan2026lavpr}. The CrossPlace re-ranker is trained on an offline hard-negative dataset mined from GSV-Cities via CrossPlace Stage~1 retrieval: for each training query, we retrieve the top-$100$ candidate images and treat all non-positive matches in this shortlist as hard negatives.

Extended training and implementation details are provided in the suppl. material.

%% file: sec/5_results.tex
\section{Experimental Results}
We first compare CrossPlace against prior work (\Cref{tab:main_results}). Guided by our motivating hypothesis, we then isolate the primary drivers of retrieval accuracy (\cref{sec:driver_performance}). We then compare our re-ranking module to alternative methods (\cref{sec:stage2_results}), analyze storage and latency tradeoffs~\cref{sec:efficiency} and ablate our core design choices~\cref{sec:ablations}.
Full Recall@5/10 results, significance tests, qualitative examples, and additional ablations are provided in the suppl.
material.

\subsection{Cross-Model Place Recognition}
\Cref{tab:main_results} reports Recall@1 across all evaluated datasets. Our Stage~1 global retrieval  outperforms LaVPR across every benchmark. On the AmsterTime, MSLS-val, and Pitts30k datasets ('Common' benchmarks), Stage~1 improves Recall@1 by $+7.8$, $+15.8$, and $+11.0$~pp, respectively. Under challenging weather and illumination conditions on SVOX, Stage~1 improves upon LaVPR by $+5.2$ to $+20.3$~pp across all settings. Similarly, on the city-scale SF-XL benchmark, Recall@1 rises from $9.5\%$ to $16.3\%$ on SF-XL-v1 ($+6.8$~pp) and from $19.6\%$ to $34.6\%$ on SF-XL-{v1} ($+15.0$~pp).

\begin{figure}[h!]
    \centering

    \begin{subfigure}[b]{0.47\linewidth}
        \centering
        \includegraphics[width=\linewidth]{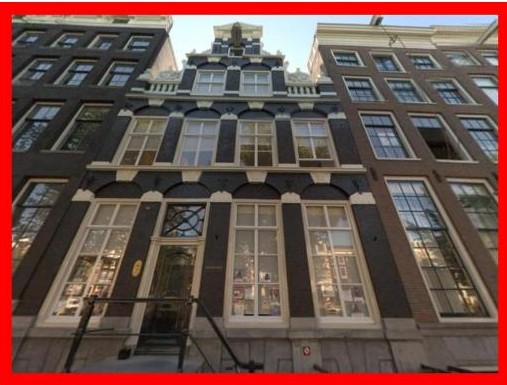}
        \caption{LaVPR}
        \label{fig:qual_lavpr_img}
    \end{subfigure}
    \hfill
    \begin{subfigure}[b]{0.47\linewidth}
        \centering
        \includegraphics[width=\linewidth]{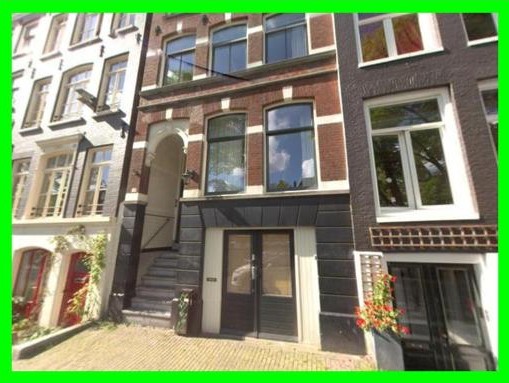}
        \caption{CrossPlace}
        \label{fig:qual_cp_img}
    \end{subfigure}

    \vspace{0.3em}

    \begin{subfigure}[b]{0.47\linewidth}
        \centering
        \includegraphics[width=\linewidth]{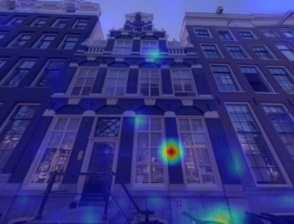}
        \caption{LaVPR gradcam}
        \label{fig:qual_lavpr_heat}
    \end{subfigure}
    \hfill
    \begin{subfigure}[b]{0.47\linewidth}
        \centering
        \includegraphics[width=\linewidth]{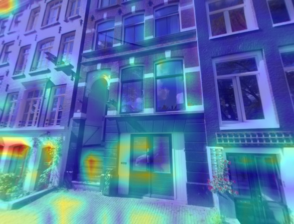}
        \caption{CrossPlace gradcam}
        \label{fig:qual_cp_heat}
    \end{subfigure}

   \caption{Qualitative comparison on an AmsterTime query (\textit{``dark double doors... arched windows...''}). LaVPR~\subref{fig:qual_lavpr_img} retrieves an incorrect candidate (red), whereas CrossPlace~\subref{fig:qual_cp_img} retrieves the correct location (green). Grad-CAM visualizations show that LaVPR activates on unrelated regions~\subref{fig:qual_lavpr_heat}, while CrossPlace focuses on the architectural landmarks described in the query~\subref{fig:qual_cp_heat}.}
    \label{fig:qualitative}
\end{figure}

Adding Stage~2 re-ranking consistently yields further performance gains. On the Common benchmarks, re-ranking improves Recall@1 to $24.0\%$, $60.1\%$, and $66.4\%$ on AmsterTime, MSLS-val, and Pitts30k, respectively, yielding overall gains of $+10.8$, $+22.1$, and $+16.2$~pp over LaVPR. Paired bootstrap tests across 1{,}000 resamples confirm that the Stage~2 gains over Stage~1 are statistically significant at all recall levels across all three datasets ($p{<}0.05$). On SVOX, CrossPlace achieves total improvements ranging from $+9.7$~pp under night illumination to $+32.4$~pp under overcast skies (from $36.8\%$ to $69.2\%$), demonstrating stability across strong photometric shifts. Similarly, on SF-XL, Stage~2 adds $+6.4$ and $+7.0$~pp over Stage~1, more than doubling Recall@1 relative to LaVPR (from $9.5\%$ to $22.7\%$ on {SF-XL-v1} and $19.6\%$ to $41.6\%$ on {SF-XL-v2}). Qualitatively (\cref{fig:qualitative}), CrossPlace attends to specific landmark cues described in the text query, whereas LaVPR focuses on uninformative background regions (illustrated through Gradient-weighted Class Activation Mapping (Grad-CAM) analysis~\cite{selvaraju2017grad}). 

\begin{table*}[t]
\centering
\caption{Recall@1 (\%) on the 'common' cross-modal benchmarks and on 
condition-specific SVOX and city-scale SF-XL splits. The last row gives the gain of 
our full pipeline over LaVPR~\cite{idan2026lavpr}. Best results in \textbf{bold}.}
\label{tab:main_results}
\small
\setlength{\tabcolsep}{5pt}
\begin{tabular}{@{}l ccc ccccc cc@{}}
\toprule
 & \multicolumn{3}{c}{Common} 
 & \multicolumn{5}{c}{SVOX} 
 & \multicolumn{2}{c}{SF-XL} \\
\cmidrule(lr){2-4} \cmidrule(lr){5-9} \cmidrule(l){10-11}
Method 
 & AmsterTime & MSLS & Pitts30k 
 & Night & Sun & Rain & Snow & Overcast 
 & v1 & v2 \\
\midrule
BLIP (zero-shot)~\cite{li2022blip} 
 & 1.3 & 1.6 & 8.1 
 & 1.5 & 0.8 & 0.5 & 1.6 & 1.6
 & 0.1 & 1.3 \\
LaVPR~\cite{idan2026lavpr}  
 & 13.2 & 38.0 & 50.2 
 & 15.7 & 30.9 & 30.7 & 29.5 & 36.8 
 & \phantom{0}9.5 & 19.6 \\
CrossPlace (Retrieval)
 & 21.0 & 53.8 & 61.2 
 & 20.9 & 45.1 & 46.7 & 49.5 & 57.1 
 & 16.3 & 34.6 \\
CrossPlace (Reranking)
 & \textbf{24.0} & \textbf{60.1} & \textbf{66.4} 
 & \textbf{25.4} & \textbf{52.0} & \textbf{59.7} & \textbf{61.5} & \textbf{69.2} 
 & \textbf{22.7} & \textbf{41.6} \\
\midrule
\textcolor{gray}{$\Delta$ vs.\ LaVPR} 
 & \textcolor{gray}{+10.8} & \textcolor{gray}{+22.1} & \textcolor{gray}{+16.2} 
 & \textcolor{gray}{+9.7} & \textcolor{gray}{+21.1} & \textcolor{gray}{+29.0} 
 & \textcolor{gray}{+32.0} & \textcolor{gray}{+32.4} 
 & \textcolor{gray}{+13.2} & \textcolor{gray}{+22.0} \\
\bottomrule
\end{tabular}
\end{table*}

\input{figures/ablations_fig}

\subsection{Drivers of Performance Gains}
\label{sec:driver_performance}
To understand what drives retrieval accuracy, we isolate the contribution of each architectural component along two main axes: visual encoder granularity and pretraining, text encoder capacity and task alignment. \Cref{fig:encoder_ablation} reports Stage~1 Recall@1 across AmsterTime, MSLS-val, and Pitts30k.

\paragraph{Visual Encoder Granularity and Pretraining.}
Visual token density is the primary driver of retrieval performance, outweighing text encoder capacity. Across every benchmark, upgrading visual granularity with the smaller text encoder (ViT-B/16 + BGE-B) outperforms upgrading text capacity with coarser patches (ViT-B/32 + BGE-L), for example, achieving $43.8\%$ vs.\ $37.8\%$ on MSLS-val and $15.4\%$ vs.\ $13.7\%$ on AmsterTime. Holding both the CLIP initialization and BGE-B fixed, increasing patch granularity from ViT-B/32 ($49$ tokens) to ViT-B/16 ($196$ tokens) isolates this effect, improving Recall@1 by $+7.7$~pp on MSLS-val (from $36.1\%$ to $43.8\%$) and $+4.4$~pp on Pitts30k (from $48.6\%$ to $53.0\%$). Switching to DINOv2-B/14 ($256$ tokens) with BGE-B, consistently raises Recall@1 further, showing that self-supervised spatial representations reinforce higher patch density even before scaling the text encoder.

\paragraph{Text Encoder Capacity.}
Scaling the text backbone from BGE-B to BGE-L yields consistent improvements across all visual encoders. Paired with DINOv2-B/14, increasing text encoder capacity raises Recall@1 from $18.1$ to $20.6\%$ on AmsterTime, $49.2\%$ to $50.3\%$ on MSLS-val, and $54.7$ to $57.8\%$ on Pitts30k.

\paragraph{Text Encoder Task Alignment.}
Progressively fine-tuning BGE-L on place descriptions ($\text{BGE-L}_{\text{ft}}$) provides further gains across every visual backbone and benchmark. When paired with DINOv2-B/14, task alignment brings Recall@1 to $21.0\%$ on AmsterTime, $53.8\%$ on MSLS-val, and $61.2\%$ on Pitts30k, outperforming the fine-tuned CLIP ViT-B/16 floor baseline by $6.7$, $10.3$, and $9.5$~pp, respectively. Across datasets, encoder capacity and domain fine-tuning provide complementary, comparable gains on the text side, with fine-tuning offering the largest boost on MSLS-val ($+3.5$~pp) and Pitts30k ($+3.4$~pp).

Taken together, these results show that all three design choices contribute additively: higher visual token density and self-supervised pretraining (DINOv2-B/14) provide the primary performance lift, while increasing text encoder capacity and aligning it to place descriptions yield steady, complementary improvements. Our full Stage~1 configuration ($\text{DINOv2-B/14} + \text{BGE-L}_{\text{ft}}$) achieves the highest accuracy across all three benchmarks.

\paragraph{Cross-Modal Alignment Gap.}
To examine whether cross-modal alignment itself remains a limiting factor, we isolate the text representation by evaluating unimodal text-to-text  retrieval, where both queries and database places are represented solely by text embeddings. Using the same fine-tuned text encoder, retrieval achieves $67.6\%$, $58.1\%$, and $22.5\%$ Recall@1 on Pitts30k, MSLS-val, and AmsterTime, respectively. CrossPlace closely matches, and on MSLS-val and AmsterTime surpasses, these unimodal reference points ($66.4\%$ vs.\ $67.6\%$ on Pitts30k, $60.1\%$ vs.\ $58.1\%$ on MSLS-val, and $24.0\%$ vs.\ $22.5\%$ on AmsterTime). Our observation that text-to-image retrieval with CrossPlace closely matches unimodal text-to-text retrieval suggests that cross-modal alignment becomes a secondary bottleneck once strong visual representations are paired with fine-grained reranking.

\begin{table*}[tbh!]
\centering
\caption{Cross-modal reranking of top-20 candidates obtained with CrossPlace Stage~1. All methods are evaluated as standalone scores ($\alpha{=}0$). 
Best results per dataset in \textbf{bold}. 
$^\dagger$Reranks the top-10 candidates only, thus R@10 is omitted.}
\label{tab:reranking}
\resizebox{\textwidth}{!}{%
\begin{tabular}{lccccccccc}
\toprule
\multirow{2}{*}{Method} &
\multicolumn{3}{c}{Amstertime} & 
\multicolumn{3}{c}{MSLS Val} & 
\multicolumn{3}{c}{Pitts30k} \\
\cmidrule(lr){2-4}\cmidrule(lr){5-7}\cmidrule(lr){8-10}
& R@1 & R@5 & R@10 & R@1 & R@5 & R@10 & R@1 & R@5 & R@10 \\
\midrule
\multicolumn{10}{l}{\textit{General-purpose cross-modal rerankers (zero-shot)}} \\
BLIP-2 ITM~\cite{li2023blip}
  & 2.4 & 14.5 & 31.4 & 17.6 & 49.3 & 69.1 & 28.6 & 68.5 & 84.4 \\
Qwen2.5-VL-72B~\cite{wu2025qwen}$^\dagger$
  & 11.5 & 36.1 & -- & 37.3 & 67.6 & -- & 48.8 & 79.9 & -- \\
\midrule
\multicolumn{10}{l}{\textit{Cross-modal rerankers for place recognition (Finetuned)}} \\
FILIP-based

  & 9.7 & 32.3 & 47.8 & 28.5 & 58.8 & 73.6 & 37.9	& 74.6 & 86.1
 \\
QFormer + ListWise Loss (baseline, ours)
   & 20.7 & 41.7 & 53.5 & 55.3 & 74.3 & 80.3 & 60.2 & 82.5 & 88.8 \\
Patch-based Cross Attention + Listwise Loss (CrossPlace, , ours)
  & \textbf{23.0} & \textbf{47.4} & \textbf{56.1}  & \textbf{57.0} & \textbf{76.2} & \textbf{81.9}  & \textbf{65.1} & \textbf{84.4} & \textbf{89.8} \\
\bottomrule
\end{tabular}%
}
\end{table*}

\subsection{Comparison with Alternative Re-Ranking Paradigms}
\label{sec:stage2_results}

To evaluate whether our token-to-patch cross-attention design is necessary, we compare it against alternative re-ranking strategies applied to the exact same Stage~1 candidate shortlists in \Cref{tab:reranking}. We benchmark our approach against learned bottleneck queries adopting the Q-Former module from~\cite{li2023blip}, token-level contrastive matching following FILIP~\cite{filip} {which applies ColBERT-style~\cite{colbert} MaxSim late interaction between text tokens and image patches}, and large zero-shot vision-language models including BLIP-2 ITM and Qwen2.5-VL-72B~\cite{wu2025qwen}.

Across all benchmarks, our token-to-patch cross-attention module achieves the highest re-ranking performance, reaching $65.1\%$ R@1 on Pitts30k, $57.0\%$ on MSLS-val, and $23.0\%$ on AmsterTime. The Q-Former baseline reaches $60.2\%$ on Pitts30k and $55.3\%$ on MSLS-val despite a comparable parameter budget, while a FILIP-style baseline~\cite{filip} with token-level contrastive supervision drops to $37.9\%$ on Pitts30k, falling below our Stage~1 CrossPlace retriever. 

Zero-shot vision-language models degrade performance even further on the shortlisted candidates. BLIP-2 ITM~\cite{li2023blip} gives $28.6\%$ R@1 on Pitts30k, while Qwen2.5-VL-72B~\cite{wu2025qwen} reaches $11.5\%$ on AmsterTime, $37.3\%$ on MSLS-val, and $48.8\%$ on Pitts30k, falling below LaVPR across all three benchmarks. Consistent with our encoder-level findings in {suppl. section G}, where a LoRA-adapted BGE-Small outperforms a frozen BGE-Large, our ${\approx}0.6$M-parameter re-ranker trained on in-distribution hard negatives outperforms an unaligned 72B model. Overall, aligning to the place recognition task on hard negatives outweighs model scale for cross-modal re-ranking.

\subsection{Storage and Latency Tradeoffs}
\label{sec:efficiency}

\Cref{tab:efficiency} compares query latency, database storage footprint, and Recall@1 on the MSLS-val dataset. Compared to LaVPR, CrossPlace Stage-1 (global retrieval) reduces query latency from $9.7$\,ms to $8.5$\,ms while increasing Recall@1 by $+15.8$~pp ($38.0\%$ to $53.8\%$) at a relatively modest storage cost of $3$\,KB per image ($56.6$\,MB total).
Adding Stage~2 re-ranking increases Recall@1 to $60.1\%$ while introducing a direct trade-off between memory footprint and runtime latency. Precomputing and caching dense patch tokens ($750$\,KB per image, or $13.4$\,GB on MSLS-val) adds only $0.5$\,ms over Stage~1, keeping total query latency at $9.0$\,ms. Conversely, computing patch tokens on-the-fly from stored $224\times224$ images ($20$\,KB per image, or $369$\,MB on MSLS-val) cuts storage requirements by over $36\times$ while requiring $17.5$\,ms per query to run $K$ DINOv2 forward passes. Because re-ranking operates strictly on the top-$K$ (=20) candidates, on-the-fly latency remains constant regardless of total database size $N$, making cached execution ideal when memory is abundant and on-the-fly execution preferable for large-scale databases.

\begin{table}[tbh!]
\centering
\footnotesize
\setlength{\tabcolsep}{4pt}
\caption{{Latency, storage, and accuracy trade-offs across retrieval and re-ranking configurations.} Evaluation is conducted on MSLS-val using $K{=}20$ shortlisted candidates, where $N$ denotes the number of database images.}
\label{tab:efficiency}
\begin{tabular}{lccc}
\toprule
\textbf{Configuration} & \textbf{Latency} & \textbf{Storage (MSLS)} & \textbf{R@1} \\
\midrule
LaVPR                    & 9.7 ms  & 1KB $\times$ N \ \ (18.9 MB) & 38.0 \\
CrossPlace Stage-1 (S1)       & 8.5 ms  & 3KB $\times$ N \ \ (56.6 MB) & 53.8 \\
\midrule
S1 + Cached reranking         & 9 ms    & 750KB $\times$ N \ (13.4 GB) & 60.1 \\
S1 + On-the-fly reranking     & 17.5 ms & 20KB $\times$ N \ \ (369 MB) & 60.1 \\
\bottomrule
\end{tabular}
\end{table}

\subsection{Ablations}
\label{sec:ablations}

\paragraph{Score fusion weight ($\alpha$, \cref{eq:score_fusion}).}
Fusing global retrieval similarity with the Stage~2 reranking score consistently outperforms reranking alone. On Pitts30k-val (\cref{fig:alpha_ablation}), adding a small retrieval weight ($\alpha{=}0.2$) improves Recall@1 from $62.9\%$ ($\alpha{=}0$) to $65.0\%$ ($+2.1$~pp), after which accuracy declines steadily toward Stage~1 retrieval. This suggests that global and local signals are complementary, and a small global weight stabilizes reranking.

\begin{figure}[t]
\centering
\begin{tikzpicture}[x=4.8cm, y=0.4cm,
    plotgrid/.style={draw=gray!25, line width=0.3pt},
    plotaxis/.style={draw=black!70, line width=0.5pt},
    serDino/.style={draw=cDino, line width=1.1pt},
    mCircle/.style={circle, fill=cDino, draw=white, line width=0.4pt,
                    inner sep=0pt, minimum size=4pt},
    mOurs/.style={circle, fill=cDino, draw=white, line width=0.6pt,
                  inner sep=0pt, minimum size=7pt}]

    \fill[cDino, opacity=0.08]
        (0.0, 0) -- (0.0, 4.9) -- (0.1, 6.6) -- (0.2, 7.0) --
        (0.3, 6.4) -- (0.4, 5.5) -- (0.5, 4.7) -- (0.6, 4.1) --
        (0.7, 3.1) -- (0.8, 1.7) -- (0.9, 1.1) -- (0.9, 0) -- cycle;

    \foreach \y/\lbl in {1/59, 3/61, 5/63, 7/65} {
        \draw[plotgrid] (0, \y) -- (1.0, \y);
        \draw[plotaxis] (0, \y) -- (-0.018, \y)
            node[left, font=\scriptsize] {\lbl};
    }

    \foreach \x/\lbl in {0.0/0.0, 0.2/0.2, 0.4/0.4, 0.6/0.6, 0.8/0.8} {
        \draw[plotaxis] (\x, 0) -- (\x, -0.18)
            node[below, font=\scriptsize] {\lbl};
    }

    \draw[cFloor16, densely dashed, line width=0.75pt]
        (0.0, 4.9) -- (1.0, 4.9);
    \node[right, font=\tiny, text=cFloor16] at (1.0, 4.9)
        {\;baseline};

    \draw[cDino, densely dashed, line width=0.65pt]
        (0.2, 0) -- (0.2, 7.0);

    \draw[serDino]
        (0.0, 4.9) -- (0.1, 6.6) -- (0.2, 7.0) -- (0.3, 6.4) --
        (0.4, 5.5) -- (0.5, 4.7) -- (0.6, 4.1) -- (0.7, 3.1) --
        (0.8, 1.7) -- (0.9, 1.1);

    \foreach \x/\y in {
        0.0/4.9, 0.1/6.6, 0.3/6.4, 0.4/5.5,
        0.5/4.7, 0.6/4.1, 0.7/3.1, 0.8/1.7, 0.9/1.1}
        \node[mCircle] at (\x, \y) {};

    \node[mOurs] at (0.2, 7.0) {};
    \node[above=3pt, font=\scriptsize\bfseries, text=cDino]
        at (0.2, 7.0) {$65.0$};
    \node[below=2pt, font=\tiny, text=cDino]
        at (0.2, 0) {$\alpha^*$};

    \draw[plotaxis] (0, 0) -- (1.03, 0);
    \draw[plotaxis] (0, 0) -- (0, 8.0);

    \node[font=\scriptsize] at (0.5, -1.2)
        {$\alpha$};
    \node[rotate=90, font=\scriptsize] at (-0.14, 4.0)
        {Recall@1 (\%)};

\end{tikzpicture}
\caption{Score fusion weight $\alpha$ on Pitts30k-val. Recall@1 peaks at $\alpha^*{=}0.2$ (\textbf{65.0\%}), outperforming the reranker-only baseline ($\alpha{=}0$, $62.9\%$, dashed line), and declines monotonically toward Stage~1 retrieval at larger $\alpha$.}
\label{fig:alpha_ablation}
\end{figure}

\paragraph{Reranker loss.}
Listwise cross-entropy outperforms binary cross-entropy (BCE) and pairwise margin loss across all benchmarks (\cref{tab:reranker_loss}). Margin loss degrades noticeably on MSLS-val ($48.5\%$ vs.\ $57.0\%$), whereas BCE remains closer ($56.9\%$). Combining listwise supervision with score fusion ($\alpha{=}0.2$) adds another $+1.3$~pp on Pitts30k and $+3.1$~pp on MSLS-val, suggesting their complementary effect.

\begin{table}[t]
\centering
\small
\caption{Reranker loss ablation (Recall@1, evaluated at $\alpha{=}0$ using the reranker score alone). We also show our selected configuration ($\alpha{=}0.2$) as a reference.}
\label{tab:reranker_loss}
\begin{tabular}{lccc}
\toprule
Configuration & AmsterTime & MSLS & Pitts30k \\
\midrule
\multicolumn{4}{l}{\textit{Selected model ($\alpha{=}0.2$)}} \\
Listwise & \textbf{24.0} & \textbf{60.1} & \textbf{66.4} \\
\midrule
\multicolumn{4}{l}{\textit{Loss comparison ($\alpha{=}0$, reranker only)}} \\
BCE      & 20.4          & 56.9          & 61.4          \\
Margin   & 19.8          & 48.5          & 60.2          \\
Listwise & \textbf{23.0} & \textbf{57.0} & \textbf{65.1} \\
\bottomrule
\end{tabular}
\end{table}

%% file: figures/ablations_fig.tex
\begin{figure*}[tbh!]
\centering

\definecolor{cClipB32}{RGB}{0, 130, 130}       
\definecolor{cClipB16}{RGB}{220, 105, 0}       
\definecolor{cDino}{RGB}{125, 40, 140}         
\definecolor{cFloor32}{RGB}{100, 100, 100}     
\definecolor{cFloor16}{RGB}{40, 90, 180}       

\tikzset{
    plotgrid/.style={gray!20, line width=0.35pt},
    plotaxis/.style={black!80, line width=0.5pt},
    floorline1/.style={cFloor32, dotted, line width=0.9pt},
    floorline2/.style={cFloor16, densely dashed, line width=0.85pt},
    serClipB32/.style={cClipB32, line width=1.0pt},
    serClipB16/.style={cClipB16, line width=1.0pt},
    serDino/.style={cDino, line width=1.0pt},
    mSquare/.style={draw=cClipB32, fill=cClipB32, regular polygon, regular polygon sides=4, inner sep=1.2pt},
    mTri/.style={draw=cClipB16, fill=cClipB16, regular polygon, regular polygon sides=3, inner sep=1.2pt},
    mCircle/.style={draw=cDino, fill=cDino, circle, inner sep=1.2pt},
    mOurs/.style={draw=cDino, fill=cDino, circle, inner sep=1.3pt, double=white, double distance=0.7pt}
}

\begin{subfigure}[b]{0.31\textwidth}
\centering
\begin{tikzpicture}[x=1.10cm, y=0.22cm]
    
    \foreach \y in {12, 14, 16, 18, 20, 22} {
        \draw[plotgrid] (0.5, \y) -- (3.5, \y);
        \draw[plotaxis] (0.5, \y) -- (0.42, \y) node[left, font=\scriptsize] {\y};
    }
    
    \draw[plotaxis] (0.5, 11) -- (0.5, 22.3);
    \draw[plotaxis] (0.5, 11) -- (3.5, 11);
    
    \node[rotate=90, font=\scriptsize] at (-0.25, 16.5) {Recall@1 (\%)};
    
    \foreach \x/\lbl in {1/{BGE-B}, 2/{BGE-L}, 3/{BGE-L$_{\text{ft}}$}} {
        \draw[plotaxis] (\x, 11) -- (\x, 10.7) node[below=1pt, font=\scriptsize] {\lbl};
    }
    
    \draw[floorline1] (0.5, 12.3) -- (3.5, 12.3);
    \draw[floorline2] (0.5, 14.3) -- (3.5, 14.3);
    
    \draw[serClipB32] (1, 12.8) -- (2, 13.7) -- (3, 14.1);
    \node[mSquare] at (1, 12.8) {};
    \node[mSquare] at (2, 13.7) {};
    \node[mSquare] at (3, 14.1) {};
    
    \draw[serClipB16] (1, 15.4) -- (2, 16.7) -- (3, 17.4);
    \node[mTri] at (1, 15.4) {};
    \node[mTri] at (2, 16.7) {};
    \node[mTri] at (3, 17.4) {};
    
    \draw[serDino] (1, 18.1) -- (2, 20.6) -- (3, 21.0);
    \node[mCircle] at (1, 18.1) {};
    \node[mCircle] at (2, 20.6) {};
    \node[mOurs] at (3, 21.0) {};
    \node[above=2pt, font=\tiny\bfseries, text=cDino] at (3, 21.0) {Ours};
\end{tikzpicture}
\caption{AmsterTime}
\label{fig:ablation_amster}
\end{subfigure}
\hfill
\begin{subfigure}[b]{0.31\textwidth}
\centering
\begin{tikzpicture}[x=1.10cm, y=0.104cm]
    
    \foreach \y in {35, 40, 45, 50, 55} {
        \draw[plotgrid] (0.5, \y) -- (3.5, \y);
        \draw[plotaxis] (0.5, \y) -- (0.42, \y) node[left, font=\scriptsize] {\y};
    }
    
    \draw[plotaxis] (0.5, 32) -- (0.5, 55.8);
    \draw[plotaxis] (0.5, 32) -- (3.5, 32);
    
    \foreach \x/\lbl in {1/{BGE-B}, 2/{BGE-L}, 3/{BGE-L$_{\text{ft}}$}} {
        \draw[plotaxis] (\x, 32) -- (\x, 31.3) node[below=1pt, font=\scriptsize] {\lbl};
    }
    
    \draw[floorline1] (0.5, 35.4) -- (3.5, 35.4);
    \draw[floorline2] (0.5, 43.5) -- (3.5, 43.5);
    
    \draw[serClipB32] (1, 36.1) -- (2, 37.8) -- (3, 38.5);
    \node[mSquare] at (1, 36.1) {};
    \node[mSquare] at (2, 37.8) {};
    \node[mSquare] at (3, 38.5) {};
    
    \draw[serClipB16] (1, 43.8) -- (2, 45.3) -- (3, 47.0);
    \node[mTri] at (1, 43.8) {};
    \node[mTri] at (2, 45.3) {};
    \node[mTri] at (3, 47.0) {};
    
    \draw[serDino] (1, 49.2) -- (2, 50.3) -- (3, 53.8);
    \node[mCircle] at (1, 49.2) {};
    \node[mCircle] at (2, 50.3) {};
    \node[mOurs] at (3, 53.8) {};
    \node[above=2pt, font=\tiny\bfseries, text=cDino] at (3, 53.8) {Ours};
\end{tikzpicture}
\caption{MSLS-val}
\label{fig:ablation_msls}
\end{subfigure}
\hfill
\begin{subfigure}[b]{0.31\textwidth}
\centering
\begin{tikzpicture}[x=1.10cm, y=0.153cm]
    
    \foreach \y in {48, 52, 56, 60} {
        \draw[plotgrid] (0.5, \y) -- (3.5, \y);
        \draw[plotaxis] (0.5, \y) -- (0.42, \y) node[left, font=\scriptsize] {\y};
    }
    
    \draw[plotaxis] (0.5, 46) -- (0.5, 62.8);
    \draw[plotaxis] (0.5, 46) -- (3.5, 46);
    
    \foreach \x/\lbl in {1/{BGE-B}, 2/{BGE-L}, 3/{BGE-L$_{\text{ft}}$}} {
        \draw[plotaxis] (\x, 46) -- (\x, 45.5) node[below=1pt, font=\scriptsize] {\lbl};
    }
    
    \draw[floorline1] (0.5, 48.7) -- (3.5, 48.7);
    \draw[floorline2] (0.5, 51.7) -- (3.5, 51.7);
    
    \draw[serClipB32] (1, 48.6) -- (2, 52.5) -- (3, 53.8);
    \node[mSquare] at (1, 48.6) {};
    \node[mSquare] at (2, 52.5) {};
    \node[mSquare] at (3, 53.8) {};
    
    \draw[serClipB16] (1, 53.0) -- (2, 56.2) -- (3, 57.7);
    \node[mTri] at (1, 53.0) {};
    \node[mTri] at (2, 56.2) {};
    \node[mTri] at (3, 57.7) {};
    
    \draw[serDino] (1, 54.7) -- (2, 57.8) -- (3, 61.2);
    \node[mCircle] at (1, 54.7) {};
    \node[mCircle] at (2, 57.8) {};
    \node[mOurs] at (3, 61.2) {};
    \node[above=2pt, font=\tiny\bfseries, text=cDino] at (3, 61.2) {Ours};
\end{tikzpicture}
\caption{Pittsburgh30k}
\label{fig:ablation_pitts}
\end{subfigure}

\vspace{0.5em}
\begin{tikzpicture}[font=\scriptsize]
    \matrix[
        matrix of nodes,
        nodes={anchor=west, inner sep=1pt},
        column sep=1.5em,
        row sep=0.35em,
        draw=gray!30,
        rounded corners=2pt,
        inner sep=4pt,
        fill=gray!5
    ] {
        \draw[serClipB32] (-0.25, 0) -- (0.25, 0); \node[mSquare] at (0, 0) {}; & \node {ViT-B/32 (CLIP init, 49 vis)}; &
        \draw[serClipB16] (-0.25, 0) -- (0.25, 0); \node[mTri] at (0, 0) {};    & \node {ViT-B/16 (CLIP init., 196 vis)}; &
        \draw[serDino] (-0.25, 0) -- (0.25, 0);    \node[mCircle] at (0, 0) {}; & \node {DINOv2-B/14, (256 vis)}; &
        \node[mOurs] at (0, 0) {};                                             & \node {\textbf{Ours (DINOv2 + BGE-L$_{\text{ft}}$)}}; \\
        \draw[floorline1] (-0.25, 0) -- (0.25, 0);                               & \node {CLIP ViT-B/32 (Floor)}; &
        \draw[floorline2] (-0.25, 0) -- (0.25, 0);                               & \node {CLIP ViT-B/16 (Floor)}; &
        & & & \\
    };
\end{tikzpicture}

\caption{Cross-modal encoder ablation across AmsterTime, MSLS-val, and Pittsburgh30k. All models (including the CLIP floor baselines) are fine-tuned using LoRA and Multi-Similarity (MS) loss. Visual encoders vary in token granularity and pretraining initialization (CLIP vs. DINOv2). Textual encoders differ by parameter scale and domain fine-tuning. DINOv2-B/14 paired with fine-tuned BGE-L$_{\text{ft}}$ achieves top performance across datasets (\textbf{Ours}).}
\label{fig:encoder_ablation}
\end{figure*}

%% file: sec/6_conclusions.tex
\section{Conclusion}
\textbf{Summary.}
We introduced CrossPlace, a two-stage cross-modal place recognition framework designed to capture fine-grained correspondences between image regions and text. Systematic analysis confirm that visual token density is a primary driver of retrieval accuracy, complemented by text encoder scale and task fine-tuning. Across existing and newly introduced benchmarks, presenting diverse environmental challenges, CrossPlace achieves double-digit Recall@1 gains over prior cross-modal baselines, closes the gap to unimodal text retrieval, and narrows it to unimodal VPR.

\textbf{Limitations and Future Work.} 
A performance gap remains relative to unimodal VPR, which may be narrowed through query-conditioned visual encoding. In addition, re-ranking tradeoffs can be mitigated through compression and quantization methods. 

%% file: sec/X_suppl.tex
This supplementary material provides additional details and results
that complement the main paper:
\begin{itemize}
    \item \Cref{sec:supp_arch} gives an overview of the full pipeline and its implementation details.
    \item \Cref{sec:supp_baselines} describes the baselines and reference systems we compare against.
    \item \Cref{sec:supp_full_results} reports full Recall@$k$ results on the common benchmarks from LaVPR~\cite{idan2026lavpr}.
    \item \Cref{sec:supp_svox_sfxl} describes how we generated text descriptions for SVOX and SF-XL and reports full Recall@$k$ results on both datasets.
    \item \Cref{sec:supp_re-rank_ablations} reports statistical significance tests for the re-ranking gains.
    \item \Cref{sec:supp_query_length} studies the impact of query length on cross-modal retrieval.
    \item \Cref{sec:supp_T2T_ablations} reports the text-to-text retrieval diagnostic.    
    \item \Cref{sec:supp_qualitative} shows additional qualitative examples.
\end{itemize}

\section{Pipeline Overview and Implementation Details}
\label{sec:supp_arch}

\begin{figure*}[!t]
    \centering
    \includegraphics[width=1.0\textwidth]{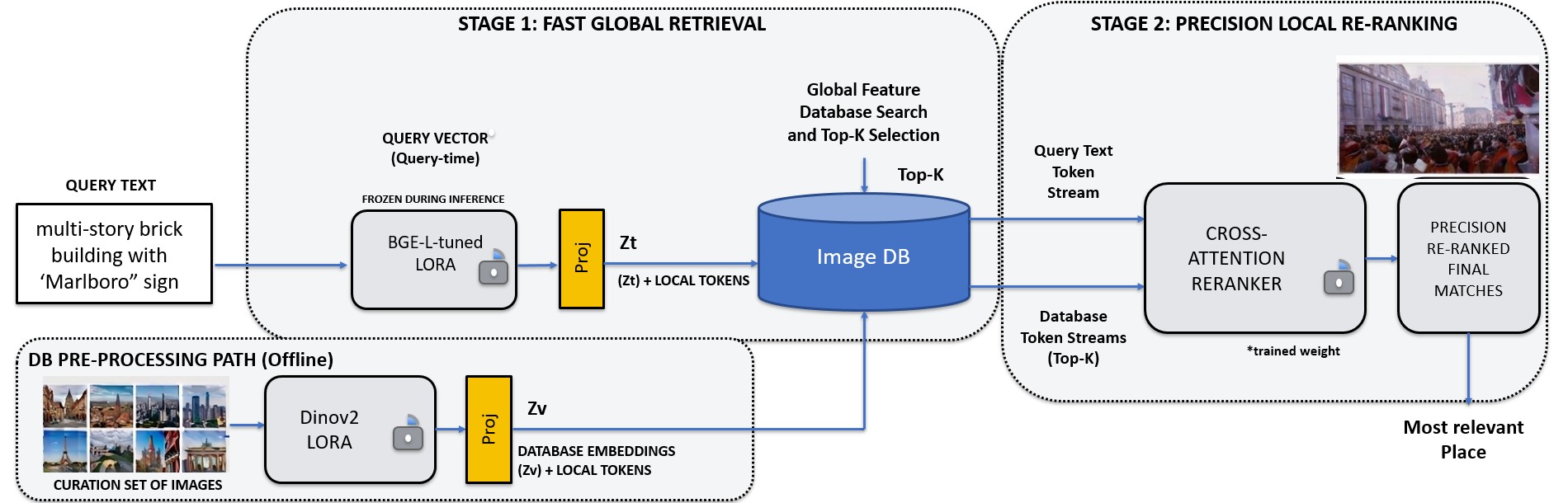}
    \caption{Overview of our two-stage pipeline. Stage~1 retrieves a
    top-$K$ shortlist via cosine similarity between LoRA-tuned BGE
    and DINOv2 global embeddings. Stage~2 re-ranks the shortlist
    with a lightweight cross-attention module operating on raw text
    tokens and image patch tokens from the frozen Stage~1 encoders,
    bypassing the projection heads, and fuses the local score with
    the global score.}
    \label{fig:supp_arch}
\end{figure*}

\Cref{fig:supp_arch} illustrates our two-stage pipeline. In Stage~1, fine-tuned
BGE-Large encodes the text query and DINOv2-B encodes the database
images, and both encoders are adapted with LoRA. Learned projection
heads map their global features into a shared 768-dim embedding
space, and the top-$K$ images by cosine similarity form a shortlist.

In Stage~2, a lightweight cross-attention re-ranker rescores each
shortlisted image. It works directly on the token-level features of
the Stage~1 encoders and bypasses the projection heads. Each text
token attends to the patch tokens of the candidate image, and the
resulting local score is fused with the Stage~1 global score. Since
only the $K$ candidates are processed, with patch tokens either
cached or recomputed on the fly, the cost of Stage~2 does not depend
on the gallery size.

\paragraph{Text queries.}
Each query is a textual description of a place. For AmsterTime,
MSLS, and Pittsburgh30k (Pitts30k) we use the descriptions released
with LaVPR~\cite{idan2026lavpr}, so that our results are directly
comparable with theirs. For SVOX and SF-XL, we generated our own
descriptions (\Cref{sec:supp_svox_sfxl}).

\paragraph{Stage 1: dual encoder.}
The text encoder, BGE-Large, is a BERT-large model with 335M
parameters. The image encoder, DINOv2-B, is a ViT-B/14 with 86M
parameters. We keep their base weights frozen and add LoRA adapters
to the attention and MLP layers of both. The database images are
encoded once, offline, so at query time we only need to encode the
text query and retrieve the top-$K$ images.

\paragraph{Stage 2: cross-attention re-ranker.}
The re-ranker works on the last-layer tokens of the Stage~1
encoders: one BGE token per word-piece, and the DINOv2 patch tokens.
Since it skips the projection heads, it can use details that the
global embeddings lose. Its interaction block
(Eqs.~(3)--(5) in the main paper) has three steps. First, each text
token attends to the image patches, so a word such as ``sign'' or
``arch'' can focus on the part of the image that shows it. Second,
self-attention between the text tokens shares this visual evidence
across the whole description. Third, a feed-forward network (hidden
size $4d{=}512$, GELU) refines each token.

The re-ranker training
uses weight decay $10^{-3}$, 200 warm-up steps, and dropout $0.1$.
For each training query, the candidate list contains the images of
its own place together with the highest-ranked images of other
places, found by running the Stage~1 model on GSV-Cities.

\section{Baselines and Reference Systems}
\label{sec:supp_baselines}

Below we describe every system that appears in the result tables:
the retrieval baselines BLIP and LaVPR, our own reference points and
ablation variants, and the re-ranking baselines of Table~2 in the
main paper.


\paragraph{BLIP (zero-shot).}
BLIP~\cite{li2022blip} is a vision--language model pretrained with
image--text contrastive, image--text matching, and captioning
objectives on large web-scale data. We use it without any
fine-tuning to measure how far a general-purpose vision--language
model gets on place retrieval without task adaptation. Its R@1 stays
below $2\%$ on every benchmark except Pitts30k ($8.1\%$), so
place-specific training is clearly needed.

\paragraph{LaVPR.}
LaVPR~\cite{idan2026lavpr} is the closest prior method for text-to-image place retrieval. The configuration we compare against adapts BLIP's text and image encoders with LoRA and trains them with a multi-similarity loss. Because we use the same adaptation and training objective, CrossPlace differs structurally in only two ways: replacing BLIP's encoders with BGE-Large and DINOv2-B/14, and adding a second stage that re-ranks the shortlist via token-level cross-attention. Critically, our contribution lies not in introducing a complex architecture, but in diagnosing what actually governs CMPR accuracy, showing that fine-grained visual token granularity, dense text capacity, and local token--patch matching resolve a primary bottleneck and narrows the performance gap to unimodal visual place recognition.

\paragraph{Text-to-text reference.}
We use text-to-text as a reference system rather than a direct cross-modal baseline. In this setting, query and
database are both text, encoded by the best-performing text-only encoder within our comparisons
(BGE-Large with all-linear LoRA, $r{=}64$, trained with the
multi-similarity loss, denoted LoRA-MS) and matched by cosine
similarity. Since no cross-modal alignment is involved, it shows how
well the text encoder alone can separate places. The gap between it
and the cross-modal systems is the alignment gap discussed in
\Cref{sec:supp_T2T_ablations}.



\paragraph{Re-ranking baselines.}
Table~2 in the main paper compares our re-ranker with four
alternatives, all applied to the same Stage~1 shortlists and
evaluated as standalone scorers ($\alpha{=}0$).

\emph{BLIP-2 ITM}~\cite{li2023blip}: we use the pretrained
image--text matching head of BLIP-2 without any fine-tuning and rank
the top-$K$ candidates by its matching score.

\emph{Qwen2.5-VL}~\cite{wu2025qwen}: we use the pretrained
Qwen2.5-VL-72B model, quantized to 8 bits, without any fine-tuning.
For computational reasons it re-ranks only the top-10 candidates.

\emph{FILIP-based}~\cite{filip}: we train a token-level
late-interaction model in the style of FILIP, with a max-similarity
(MaxSim) contrastive loss between the text tokens and the image
patch tokens. The original FILIP model was trained with a very large
batch (65{,}536) on roughly 300M image--text pairs from many
domains. Our training set, GSV-Cities, contains only about 0.5M
image--text pairs from a single urban domain, and we train with a
much smaller batch. We attribute the low accuracy of this baseline
to these differences, since token-level contrastive learning relies
on many in-batch negatives and large-scale data.

\emph{Q-Former}: a set of learned bottleneck queries, following the
Q-Former design of BLIP-2~\cite{li2023blip}, is trained on top of
the frozen CrossPlace Stage~1 encoders. The queries attend to the
image patch tokens and to the text tokens, and their pooled output is
mapped to a scalar score. It is trained with the same listwise loss
and the same hard-negative shortlists as our re-ranker.

\section{Full Results on the Common Benchmarks}
\label{sec:supp_full_results}

Supplemenrary \Cref{tab:supp_full_recall} extends Table~1 of the main paper with
R@5 and R@10 results on the three common benchmarks, and adds the text-to-text
reference row for comparison with the cross-modal methods.
CrossPlace outperforms cross-modal baselines across datasets and R@$k\in\{1,5,10\}$.
\begin{table*}[!t]
\centering
\scriptsize
\caption{Full Recall@$k$ (\%) results on the common benchmarks. The
Text-to-Text reference uses text embeddings for both query and database.
Best cross-modal results in \textbf{bold}.}
\label{tab:supp_full_recall}
\small
\setlength{\tabcolsep}{4.5pt}
\begin{tabular}{@{}l ccc ccc ccc@{}}
\toprule
 & \multicolumn{3}{c}{AmsterTime}
 & \multicolumn{3}{c}{MSLS-val}
 & \multicolumn{3}{c}{Pitts30k} \\
\cmidrule(lr){2-4}\cmidrule(lr){5-7}\cmidrule(l){8-10}
Method
 & R@1 & R@5 & R@10
 & R@1 & R@5 & R@10
 & R@1 & R@5 & R@10 \\
\midrule
BLIP (zero-shot)~\cite{li2022blip}
 & 1.3 & 4.1 & 7.0
 & 1.6 & 5.9 & 8.2
 & 8.1 & 26.8 & 41.1 \\
LaVPR~\cite{idan2026lavpr}
 & 13.2 & 31.8 & 42.3
 & 38.0 & 60.5 & 68.4
 & 50.2 & 74.2 & 82.8 \\
CrossPlace (Retrieval)
 & 21.0 & 44.5 & 53.9
 & 53.8 & 72.4 & 79.6
 & 61.2 & 81.2 & 87.8 \\
CrossPlace (Reranking)
 & \textbf{24.0} & \textbf{48.5} & \textbf{57.2}
 & \textbf{60.1} & \textbf{76.8} & \textbf{81.9}
 & \textbf{66.4} & \textbf{84.4} & \textbf{89.6} \\
 \midrule \midrule
\textit{Text-to-Text reference}
 & \textit{22.5} & \textit{42.1} & \textit{51.7}
 & \textit{58.1} & \textit{75.8} & \textit{82.7}
 & \textit{67.6} & \textit{86.9} & \textit{92.0} \\
\bottomrule
\end{tabular}
\end{table*}
\section{SVOX and SF-XL: Text Descriptions and Full Results}
\label{sec:supp_svox_sfxl}

\paragraph{Text descriptions.}
SVOX and SF-XL were not annotated in LaVPR~\cite{idan2026lavpr}, so
we generated their text descriptions ourselves with Gemini~2.5
Flash, following the captioning protocol of LaVPR, whose pipeline
was verified in that work.

\paragraph{Full results.}
\Cref{tab:svox_results} reports R@1, R@5, and R@10 on SVOX across
its five imaging conditions, and \Cref{tab:sfxl_results} reports
the same metrics on the two SF-XL query sets. These tables extends the results shown in the main text.

\begin{table*}[!t]
\centering
\small
\caption{Recall@$k$ (\%) on SVOX across five imaging conditions. Best results per
column are highlighted in \textbf{bold}.}
\label{tab:svox_results}
\setlength{\tabcolsep}{2pt}
\begin{tabular}{@{}l ccc ccc ccc ccc ccc@{}}
\toprule
 & \multicolumn{3}{c}{Night}
 & \multicolumn{3}{c}{Sun}
 & \multicolumn{3}{c}{Rain}
 & \multicolumn{3}{c}{Snow}
 & \multicolumn{3}{c}{Overcast} \\
\cmidrule(lr){2-4} \cmidrule(lr){5-7} \cmidrule(lr){8-10}
\cmidrule(lr){11-13} \cmidrule(l){14-16}
Method
 & R@1 & R@5 & R@10
 & R@1 & R@5 & R@10
 & R@1 & R@5 & R@10
 & R@1 & R@5 & R@10
 & R@1 & R@5 & R@10 \\
\midrule
BLIP (zero-shot)~\cite{li2022blip}
 & 1.5 & 5.0 & 7.7
 & 0.8 & 4.0 & 6.6
 & 0.5 & 4.7 & 6.7
 & 1.6 & 4.5 & 6.4
 & 1.6 & 3.8 & 6.0 \\
LaVPR~\cite{idan2026lavpr}
 & 15.7 & 31.7 & 41.3
 & 30.9 & 52.7 & 63.3
 & 30.7 & 52.6 & 63.7
 & 29.5 & 53.6 & 64.7
 & 36.8 & 61.1 & 71.4 \\
CrossPlace (Retrieval)
 & 20.9 & 38.0 & 48.4
 & 45.1 & 68.9 & 78.0
 & 46.7 & 70.0 & 79.3
 & 49.5 & 73.1 & 80.8
 & 57.1 & 81.7 & 88.4 \\
CrossPlace (Reranking)
 & \textbf{25.4} & \textbf{42.3} & \textbf{51.4}
 & \textbf{52.0} & \textbf{74.2} & \textbf{80.2}
 & \textbf{59.7} & \textbf{77.7} & \textbf{84.0}
 & \textbf{61.5} & \textbf{79.2} & \textbf{85.4}
 & \textbf{69.2} & \textbf{85.1} & \textbf{90.4} \\
\bottomrule
\end{tabular}
\end{table*}
\begin{table*}[!t]
\centering
\small
\caption{Recall@$k$ (\%) on SF-XL~\cite{berton2022rethinking} for the
two query sets (v1 and v2 in the main paper). Best results per
column are highlighted in \textbf{bold}.}
\label{tab:sfxl_results}
\begin{tabular}{l ccc ccc}
\toprule
\multirow{2}{*}{Method} &
\multicolumn{3}{c}{Queries-v1} &
\multicolumn{3}{c}{Queries-v2} \\
\cmidrule(lr){2-4}\cmidrule(lr){5-7}
& R@1 & R@5 & R@10
  & R@1 & R@5 & R@10 \\
\midrule
BLIP (zero-shot)~\cite{li2022blip}
  & 0.1  & 0.6  & 1.3
  & 1.3  & 3.2  & 4.5 \\
LaVPR~\cite{idan2026lavpr}
  & 9.5  & 19.9 & 26.4
  & 19.6 & 37.0 & 46.7 \\
CrossPlace (Retrieval)
  & 16.3 & 31.5 & 39.6
  & 34.6 & 54.0 & 63.7 \\
CrossPlace (Reranking)
  & \textbf{22.7} & \textbf{36.7} & \textbf{43.8}
  & \textbf{41.6} & \textbf{61.5} & \textbf{67.6} \\
\bottomrule
\end{tabular}
\end{table*}

\section{Re-ranking Significance Analysis}
\label{sec:supp_re-rank_ablations}

\Cref{tab:supp_bootstrap} reports 95\% confidence intervals from a
paired bootstrap over the test queries (1{,}000 resamples). Since
both stages are evaluated on the same resampled queries, the interval
on their difference accounts for the correlation between them.
$\Delta$ is computed from unrounded values, so it may differ by $0.1$
from the difference of the rounded numbers in
\Cref{tab:supp_full_recall}. Re-ranking improves significantly over
Stage~1 at R@1, R@5, and R@10 on all three benchmarks
($p\le 0.006$). The gain is usually largest at R@1, reaching
$+6.4$~pp on MSLS-val, and shrinks as $k$ grows. On AmsterTime it
peaks at R@5. The last column counts queries whose outcome changes:
re-ranking fixes $1.5$--$3.1\times$ more Stage~1 errors than it
introduces.

\begin{table*}[!t]
\centering
\caption{Paired bootstrap analysis of CrossPlace Stage~2 re-ranking vs.\
CrossPlace Stage~1 retrieval (1{,}000 resamples, 95\% CIs). ``Fixed / broken''
counts queries that are correct only after re-ranking / only before.} 
\label{tab:supp_bootstrap}
\begin{tabular}{llccccc}
\toprule
Dataset & Metric & Stage~1 & Stage~2& $\Delta$ & $p$ & Fixed / broken \\
\midrule
\multirow{3}{*}{AmsterTime}
  & R@1  & 21.0 [18.8, 23.4] & 24.0 [21.6, 26.4] & +2.9 [+1.0, +4.9] & 0.006    & 102 / 66 \\
  & R@5  & 44.5 [41.8, 47.1] & 48.5 [45.5, 51.3] & +4.0 [+1.9, +6.2] & $<$0.001 & 110 / 61 \\
  & R@10 & 53.9 [51.2, 56.8] & 57.2 [54.2, 60.0] & +3.2 [+1.6, +4.8] & $<$0.001 &  73 / 33 \\
\midrule
\multirow{3}{*}{MSLS-val}
  & R@1  & 53.8 [50.1, 57.0] & 60.1 [56.8, 63.4] & +6.4 [+4.1, +9.1] & $<$0.001 &  73 / 26 \\
  & R@5  & 72.4 [69.2, 75.5] & 76.8 [73.6, 79.6] & +4.3 [+2.2, +6.4] & $<$0.001 &  47 / 15 \\
  & R@10 & 79.6 [76.6, 82.4] & 81.9 [79.1, 84.6] & +2.3 [+0.8, +3.9] & 0.002    &  27 / 10 \\
\midrule
\multirow{3}{*}{Pitts30k}
  & R@1  & 61.2 [60.1, 62.4] & 66.4 [65.3, 67.5] & +5.2 [+4.3, +6.1] & $<$0.001 & 689 / 335 \\
  & R@5  & 81.2 [80.3, 82.1] & 84.4 [83.5, 85.3] & +3.2 [+2.5, +3.9] & $<$0.001 & 410 / 192 \\
  & R@10 & 87.8 [87.0, 88.5] & 89.6 [88.8, 90.2] & +1.7 [+1.2, +2.2] & $<$0.001 & 224 / 105 \\
\bottomrule
\end{tabular}
\end{table*}

\section{Impact of Query Length on Cross-Modal Retrieval}
\label{sec:supp_query_length}

Our main experiments use the full text descriptions as queries. To
study how sensitive cross-modal retrieval is to query length, we
created shorter versions of the queries: each full description was
rewritten with Claude Opus~5.5 into a shorter, more natural text of
about 40 words that keeps only its most discriminative content. Only
the queries are shortened; the image database is unchanged. We
evaluate on AmsterTime and MSLS-val. \Cref{tab:supp_query_length}
reports the 40-word results, and the full-description results are
given in \Cref{tab:supp_full_recall}.

Shorter queries reduce accuracy for all methods, showing that richer
descriptions carry useful information for localization.
Nevertheless, CrossPlace still clearly outperforms LaVPR with 40-word
queries, and Stage~2 re-ranking still improves over Stage~1
retrieval.

\begin{table*}[!t]
\centering
\caption{Recall@$k$ (\%) with 40-word queries on AmsterTime and
MSLS-val. Best results in \textbf{bold}.}
\label{tab:supp_query_length}
\small
\setlength{\tabcolsep}{4.5pt}
\begin{tabular}{@{}l ccc ccc@{}}
\toprule
 & \multicolumn{3}{c}{AmsterTime (40 words)}
 & \multicolumn{3}{c}{MSLS-val (40 words)} \\
\cmidrule(lr){2-4}\cmidrule(l){5-7}
Method
 & R@1 & R@5 & R@10
 & R@1 & R@5 & R@10 \\
\midrule
LaVPR~\cite{idan2026lavpr}
 & 10.7 & 27.9 & 36.3
 & 27.4 & 47.8 & 54.9 \\
CrossPlace (Retrieval)
 & 16.2 & 34.3 & 44.6
 & 36.9 & 55.0 & 61.5 \\
CrossPlace (Reranking)
 & \textbf{17.4} & \textbf{37.0} & \textbf{45.9}
 & \textbf{42.4} & \textbf{59.2} & \textbf{64.6} \\
\bottomrule
\end{tabular}
\end{table*}

\section{Text-to-Text Retrieval Diagnostic}
\label{sec:supp_T2T_ablations}

\Cref{tab:t2t} isolates the effect of the modality gap by comparing cross-modal (text-to-image) retrieval against a uni-modal (text-to-text) diagnostic reference in which both queries and database entries are textual descriptions. Within the uni-modal experiments, task adaptation outweighs model scale. On Pitts30k, full LoRA adaptation lifts BGE-Large from $36.5\%$ (frozen) to $67.6\%$ Recall@1, whereas Learned Language Pooling (LLP)~\cite{idan2026lavpr} alone reaches only $46.8\%$. Adapting all linear layers also proves more effective than increasing rank: all-linear LoRA at $r{=}16$ ($64.7\%$) outperforms QV LoRA at $r{=}64$ ($63.2\%$). Furthermore, a fully adapted BGE-Small ($61.5\%$ R@1), despite being roughly $10\times$ smaller, substantially outperforms frozen BGE-Large ($36.5\%$) and even surpasses BGE-Large adapted with QV LoRA at $r{=}16$ ($57.5\%$). Across all three datasets, the strongest uni-modal configuration (BGE-Large with all-linear LoRA at $r{=}64$) establishes a text-only reference of $22.5\%$, $58.1\%$, and $67.6\%$ R@1 on AmsterTime, MSLS-val, and Pitts30k, respectively, with substantial gains compared to the frozen and LLP configurations.

Comparing the cross-modal block against this uni-modal reference shows how effectively CrossPlace bridges the modality gap. While LaVPR~\cite{idan2026lavpr} trails the best text-to-text model by $9.3$, $20.1$, and $17.4$~pp Recall@1 on Amstertime, MSLS-val and Pitts30k, respectively, CrossPlace closes most of this margin in Stage~1 ($21.0\%$, $53.8\%$, and $61.2\%$ R@1) and eliminates it with Stage~2 re-ranking, surpassing the text-to-text reference on AmsterTime ($24.0\%$ vs.\ $22.5\%$) and MSLS-val ($60.1\%$ vs.\ $58.1\%$) while closely matching it on Pitts30k ($66.4\%$ vs.\ $67.6\%$). 
CrossPlace's near-parity with pure text-to-text matching suggests that when combining dense visual tokens, strong adapted text representations, and local cross-modal matching, cross-modal alignment becomes a secondary limiting factor rather than the primary bottleneck.
\begin{table*}[!t]
\centering
\caption{Text-to-text retrieval diagnostic. In the uni-modal rows,
both query and database entries are textual descriptions, which
removes cross-modal alignment entirely. LoRA adaptation matters more
than model size, and the best text-to-text result (BGE-Large,
LoRA-MS All $r{=}64$) serves as the reference for cross-modal
(text-to-image) retrieval. Cross-modal scores that exceed this
reference are shown in \textcolor{blue}{blue}. LoRA-MS denotes LoRA
fine-tuning with the multi-similarity loss, applied either to the
query and value projections (QV) or to all linear layers (All). LLP
denotes Learned Language Pooling~\cite{idan2026lavpr}.}
\label{tab:t2t}
\resizebox{\textwidth}{!}{%
\begin{tabular}{ll c ccc ccc ccc}
\toprule
\multirow{2}{*}{Model} & \multirow{2}{*}{Method} &
\multirow{2}{*}{Dim} &
\multicolumn{3}{c}{AmsterTime} &
\multicolumn{3}{c}{MSLS-val} &
\multicolumn{3}{c}{Pitts30k} \\
\cmidrule(lr){4-6}\cmidrule(lr){7-9}\cmidrule(lr){10-12}
& & & R@1 & R@5 & R@10
    & R@1 & R@5 & R@10
    & R@1 & R@5 & R@10 \\
\midrule
\multicolumn{12}{l}{\small\textit{Cross-modal (text-to-image)}} \\
LaVPR~\cite{idan2026lavpr} & LoRA-MS All $r{=}64$ & 256
 & 13.2 & 31.8 & 42.3
 & 38.0 & 60.5 & 68.4
 & 50.2 & 74.2 & 82.8 \\
CrossPlace (Retrieval) & LoRA-MS All $r{=}64$ & 768
 & 21.0 & \textcolor{blue}{44.5} & \textcolor{blue}{53.9}
 & 53.8 & 72.4 & 79.6
 & 61.2 & 81.2 & 87.8 \\
CrossPlace (Reranking) & LoRA-MS All $r{=}64$ & 768
 & \textcolor{blue}{\textbf{24.0}} & \textcolor{blue}{\textbf{48.5}} & \textcolor{blue}{\textbf{57.2}}
 & \textcolor{blue}{\textbf{60.1}} & \textcolor{blue}{\textbf{76.8}} & 81.9
 & 66.4 & 84.4 & 89.6 \\
\midrule \midrule
\multicolumn{12}{l}{\small\textit{Uni-modal (text-to-text)}} \\
BGE-Large & Frozen              & 1024
  &  9.5 & 18.0 & 23.5
  & 26.5 & 43.9 & 51.5
  & 36.5 & 62.4 & 73.3 \\
BGE-Large & LLP                 & 1024
  & 12.9 & 25.8 & 34.8
  & 37.6 & 57.7 & 66.6
  & 46.8 & 72.7 & 82.0 \\
BGE-Large & LoRA-MS QV $r{=}16$ & 1024
  & 19.3 & 36.8 & 44.9
  & 46.1 & 69.7 & 76.6
  & 57.5 & 81.8 & 88.5 \\
BGE-Large & LoRA-MS QV $r{=}64$ & 1024
  & 21.4 & 39.4 & 48.1
  & 54.1 & 73.8 & 80.7
  & 63.2 & 84.7 & 90.3 \\
BGE-Large & LoRA-MS All $r{=}16$ & 1024
  & 22.2 & 40.9 & 51.0
  & 55.5 & 74.9 & 81.6
  & 64.7 & 85.5 & 90.9 \\
BGE-Large & LoRA-MS All $r{=}64$ & 1024
  & \textbf{22.5} & \textbf{42.1} & \textbf{51.7}
  & \textbf{58.1} & \textbf{75.8} & \textbf{82.7}
  & \textbf{67.6} & \textbf{86.9} & \textbf{92.0} \\
\cmidrule(l){1-12}
BGE-Small & Frozen               & 384
  &  7.9 & 17.5 & 23.1
  & 23.4 & 43.2 & 50.3
  & 34.0 & 60.2 & 71.5 \\
BGE-Small & LoRA-MS QV $r{=}16$  & 384
  & 14.8 & 30.5 & 37.5
  & 41.4 & 62.4 & 69.3
  & 50.6 & 75.2 & 83.1 \\
BGE-Small & LoRA-MS All $r{=}64$ & 384
  & 19.2 & 38.0 & 46.6
  & 52.0 & 74.2 & 80.0
  & 61.5 & 83.3 & 90.2 \\
\cmidrule(l){1-12}
SBERT-Small & Frozen               & 384
  &  5.4 & 11.8 & 16.2
  & 17.4 & 31.9 & 38.8
  & 29.0 & 55.7 & 67.5 \\
SBERT-Small & LoRA-MS QV $r{=}16$  & 384
  & 10.6 & 24.4 & 33.0
  & 35.1 & 59.1 & 67.0
  & 46.7 & 72.2 & 81.7 \\
SBERT-Small & LoRA-MS All $r{=}64$ & 384
  & 18.5 & 35.3 & 44.8
  & 49.2 & 70.4 & 78.0
  & 58.4 & 81.5 & 88.5 \\
\bottomrule
\end{tabular}%
}
\end{table*}

\section{Qualitative Results}
\label{sec:supp_qualitative}
\begin{figure*}[!t]
    \centering

    \begin{subfigure}[c]{0.34\textwidth}
        \footnotesize
        \raggedright
        Partially visible building with a \textbf{gabled roof} on the
        left, large multi-story commercial building with a dark facade,
        numerous rectangular windows, and decorative elements, prominent
        central gable, mansard roof with dormer windows,
        ``\textbf{PETRUS CLOPPENBORG}'' signage on the roof, repeating
        decorative balustrades below upper-floor windows,
        ``REGENJAS'' signage on the ground floor, large display windows
        and arched entrances on the ground floor, wide open street or
        square in the foreground, side facade of the main building
        extending to the right.
        \caption{Query text}
        \label{fig:q963_text}
    \end{subfigure}
    \hfill
    \begin{subfigure}[c]{0.31\textwidth}
        \centering
        \includegraphics[width=\linewidth]{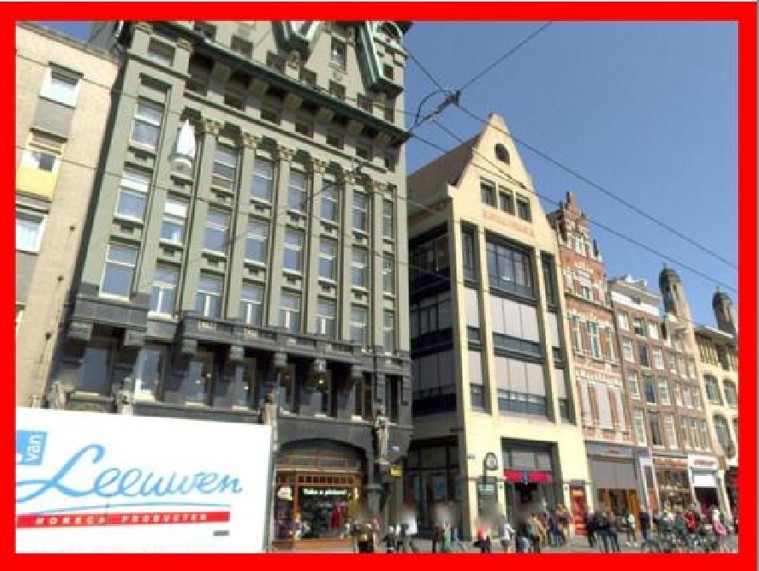}
        \caption{LaVPR}
        \label{fig:q963_lavpr}
    \end{subfigure}
    \hfill
    \begin{subfigure}[c]{0.31\textwidth}
        \centering
        \includegraphics[width=\linewidth]{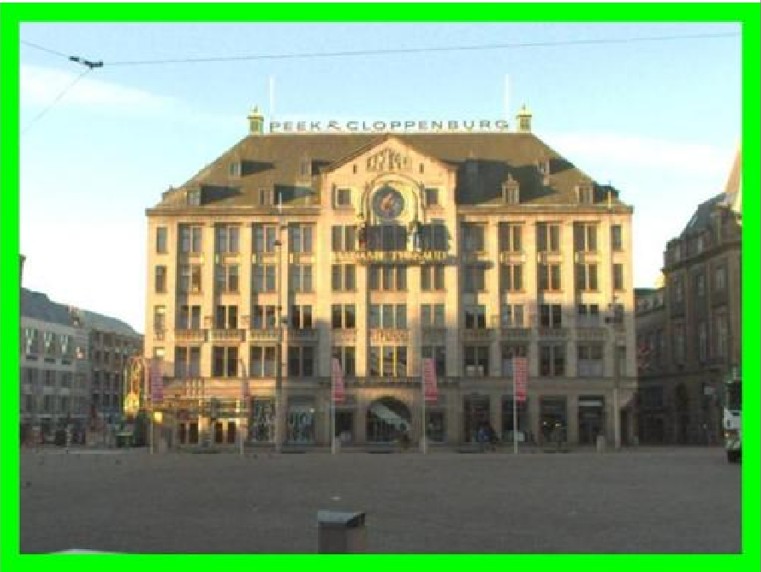}
        \caption{CrossPlace Retrieval}
        \label{fig:q963_cp}
    \end{subfigure}

    \caption{Qualitative R@1 comparison between LaVPR and CrossPlace
    Stage~1 retrieval. For the query text~\subref{fig:q963_text},
    LaVPR~\subref{fig:q963_lavpr} retrieves an incorrect place, whereas
    CrossPlace~\subref{fig:q963_cp} retrieves the correct location,
    matching distinctive cues such as the ``PETRUS CLOPPENBORG''
    signage.}
    \label{fig:supp_qualitative}
\end{figure*}

\begin{figure*}[!t]
    \centering

    \begin{subfigure}[c]{0.30\textwidth}
        \scriptsize
        \raggedright
        Part of dark building facade, dark lower section of main
        building with white vertical trim, gray stone steps, white
        arched entrance frame, dark main entrance door with `11' number
        plate, red brick facade with white horizontal bands and vertical
        accents, \textbf{ground-level dark double doors with white
        frame, two arched windows on first floor, two arched windows on
        second floor, three arched windows on third floor}, part of dark
        building facade, patterned street pavement, two dark bollards.
        \caption{Query text}
        \label{fig:q1171_text}
    \end{subfigure}
    \hfill
    \begin{subfigure}[c]{0.22\textwidth}
        \centering
        \includegraphics[width=\linewidth]{figures/1171_lavpr.jpg}
        \caption{LaVPR}
        \label{fig:q1171_lavpr}
    \end{subfigure}
    \hfill
    \begin{subfigure}[c]{0.22\textwidth}
        \centering
        \includegraphics[width=\linewidth]{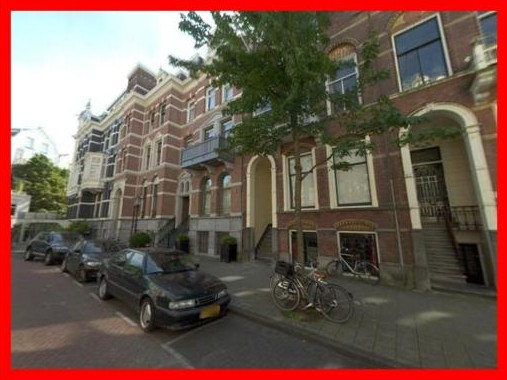}
        \caption{CrossPlace Retrieval}
        \label{fig:q1171_s1}
    \end{subfigure}
    \hfill
    \begin{subfigure}[c]{0.22\textwidth}
        \centering
        \includegraphics[width=\linewidth]{figures/1171_s2.jpg}
        \caption{CrossPlace Reranking}
        \label{fig:q1171_s2}
    \end{subfigure}

    \caption{Qualitative R@1 comparison between LaVPR, CrossPlace
    Stage~1 retrieval, and CrossPlace Stage~2 re-ranking. For the query
    text~\subref{fig:q1171_text}, both LaVPR~\subref{fig:q1171_lavpr}
    and Stage~1 retrieval~\subref{fig:q1171_s1} return incorrect
    places, whereas Stage~2 re-ranking~\subref{fig:q1171_s2} recovers
    the correct location by matching fine-grained cues such as the
    dark double doors and the arched windows.}
    \label{fig:supp_qualitative2}
\end{figure*}

\Cref{fig:supp_qualitative,fig:supp_qualitative2} show additional
qualitative comparisons of the top-1 retrieved image for a text query.
\Cref{fig:supp_qualitative} compares LaVPR with our Stage~1 retrieval,
and \Cref{fig:supp_qualitative2} adds our Stage~2 re-ranking, showing
how the retrieved place changes from the baseline to the full
pipeline.

